\documentclass[pdflatex,sn-apa]{sn-jnl}

\usepackage{graphicx}%
\usepackage{multirow}%
\usepackage{amsmath,amssymb,amsfonts}%
\usepackage{amsthm}%
\usepackage{mathrsfs}%
\usepackage[title]{appendix}%
\usepackage{xcolor}%
\usepackage{textcomp}%
\usepackage{manyfoot}%
\usepackage{booktabs}%
\usepackage{algorithm}%
\usepackage{algorithmicx}%
\usepackage{algpseudocode}%
\usepackage{listings}%
\usepackage{makecell}%

\theoremstyle{thmstyleone}%
\theoremstyle{thmstyletwo}%

\theoremstyle{thmstylethree}%

\begin{document}

\title[Benchmarking Adversarial Robustness and Adversarial Training Strategies for Object Detection]{Benchmarking Adversarial Robustness and Adversarial Training Strategies for Object Detection}


\author*[1]{\fnm{Alexis} \sur{Winter}}\email{alexis.winter@cea.fr}
\equalcont{These authors contributed equally to this work.}

\author[1]{\fnm{Jean-Vincent} \sur{Martini}}
\equalcont{These authors contributed equally to this work.}

\author[1]{\fnm{Romaric} \sur{Audigier}}

\author[1]{\fnm{Angelique} \sur{Loesch}}

\author[1]{\fnm{Bertrand} \sur{Luvison}}

\affil*[1]{\orgname{Université Paris-Saclay, CEA}, \orgdiv{List}, \postcode{F-91191}, \city{Palaiseau}, \country{France}}

\abstract{Object detection models are critical components of automated systems, such as autonomous vehicles and perception-based robots, but their sensitivity to adversarial attacks poses a serious security risk. Progress in defending these models lags behind classification, hindered by a lack of standardized evaluation. It is nearly impossible to thoroughly compare attack or defense methods, as existing work uses different datasets, inconsistent efficiency metrics, and varied measures of perturbation cost. This paper addresses this gap by investigating three key questions: (1) How can we create a fair benchmark to impartially compare attacks? (2) How well do modern attacks transfer across different architectures, especially from Convolutional Neural Networks to Vision Transformers? (3) What is the most effective adversarial training strategy for robust defense? To answer these, we first propose a unified benchmark framework focused on digital, non-patch-based attacks. This framework introduces specific metrics to disentangle localization and classification errors and evaluates attack cost using multiple perceptual metrics. Using this benchmark, we conduct extensive experiments on state-of-the-art attacks and a wide range of detectors. Our findings reveal two major conclusions: first, modern adversarial attacks against object detection models show a significant lack of transferability to transformer-based architectures. Second, we demonstrate that the most robust adversarial training strategy leverages a dataset composed of a mix of high-perturbation attacks with different objectives (e.g., spatial and semantic), which outperforms training on any single attack.
}

\keywords{Object Detection, Adversarial Robustness, Adversarial Attacks, Benchmark}



\maketitle

\paragraph{Acknowledgments}

Funded by the European Union. Views and opinions expressed are however those of the author(s) only and do not necessarily reflect those of the European Union nor the European Commission. Neither the European Union nor the granting authority can be held responsible for them. This work was supported under the EDF Project FaRADAI (grant number 101103386).
This work was made possible by the FactoryIA supercomputer (funded by the Ile-de-France Regional Council).
This work was supported by a State grant managed by the French National Research Agency, Agence Nationale de la Recherche, under "France 2030" (grant reference "ANR-22-PECY-0011", within the framework of the "COMPROMIS" project).

\newpage

\section{Introduction}

Object detectors are central to the deployment of vision systems in open-world environments, yet ensuring their trustworthiness remains a critical challenge as they remain highly vulnerable to adversarial attacks \citep{FirstSurvey, BestSurvey}.
In machine learning, adversarial samples are inputs formed by applying specific perturbations to a clean sample that cause a model to produce an incorrect output with high confidence \citep{goodfellow}. In object detection, adversarial attacks generate perturbations that alter the input images, making detectors ineffective and leading to incorrect predictions which pose a major security threat in applications like autonomous driving, video surveillance, and medical imaging \citep{FirstSurvey, BestSurvey}. Understanding such attacks and improving the robustness of our detectors has therefore become a key research topic to overcome this challenge.

While adversarial robustness has been well studied for classification in recent years \citep{ClassifSurvey, ClassifSurveyCAAI, ClassifSurveyDefense}, progress in other domains such as deep metric learning \citep{Metric1, ReID, Metric3} and object detection is far more limited. This lag is due to two primary issues: the task's inherent complexity and a lack of standardized evaluation. Unlike classification, where an attack's success is a binary outcome, an attack on an object detector can succeed in multiple ways: it can cause an object to go undetected, change its predicted class label, or alter the coordinates of its bounding box \citep{ShiftAtt, Pick-Object, OSFD}. This multi-objective failure space makes both designing attacks and evaluating defenses a challenging task. A wide variety of attack strategies have emerged for all these different objectives, from imperceptible perturbations that are invisible to the human eye \citep{EBAD, CAA}, to physical patch attacks that add a visible artifact to the scene \citep{T-SEA, DPatch}.  More critically, the field lacks any standardized evaluation. Existing surveys \citep{FirstSurvey} provide excellent taxonomies and recent research like \citep{BestSurvey} attempts a large-scale comparison, but their work highlights, the difficulty of comparing different attack or defense methodologies and the need for standard evaluation. Even among digital, non-patch-based attacks, existing work uses different datasets and efficiency metrics aligned with their respective objectives (e.g., mAP drop vs. attack success rate). Perturbation constraints are not applied in the same way even though most methods use Lebesgue norms \citep{CAA, OSFD}, and methods often require varying degrees of information regarding the target model from full white-box access \citep{OSFD} to query-based grey-box settings \citep{EBAD}. As detailed in Section \ref{sec:criticsurvey}, this fragmentation makes it impossible to assess the true state of the art or identify which methods are genuinely robust.

To address these gaps, we first introduce a critical survey and a new unified benchmark framework for digital, non-patch-based attacks. We then conduct extensive experiments on SOTA attacks, a wide range of detectors, and multiple adversarial training configurations. Our main contributions are as follows:

\begin{itemize}
    \item We analyze the landscape of adversarial attacks to identify and categorize the most effective methodologies. This analysis exposes the fragmentation of the field, highlighting the disparate datasets, metrics, and perturbation constraints that make comparisons impossible.
    \item Based on our findings, we propose a unified benchmark focused on digital, non-patch-based attacks. This framework uses specialized metrics ($\text{AP}_{\text{loc}}$ and CSR) to disentangle localization and classification errors, and two perceptual metrics (LPIPS, $L_2$) to enable fair comparisons of attack efficiency and transferability.
    \item We provide an analysis of cross-architectural transferability, revealing a robustness gap where modern attacks fail to transfer to modern transformer-based detectors like DINO.
    \item We demonstrate that training on a mix of high-perturbation attacks with complementary objectives (e.g., spatial and semantic) achieves greater robustness than training on any single attack.
\end{itemize}

\section{Landscape Analysis: Taxonomy, Method Review and Evaluation Gaps}
\label{sec:survey}

\subsection{Preliminaries and terminology}

\subsubsection{Formal definition of adversarial examples}

Adversarial examples are crafted to fool Deep Neural Networks by introducing small perturbations that lead to incorrect predictions. Formally, given a trained model with parameters $\theta$, an input image $x$, and its ground-truth label $y$, the standard training objective is to find parameters $\theta$ that \textit{minimize} a loss function $\mathcal{L}(x, \theta, y)$. In contrast, an adversarial attack seeks to find a small perturbation $\delta$ that \textit{maximizes} this loss, thereby inducing a wrong prediction:
$$
\max_{\delta \in \Delta} \mathcal{L}(x + \delta, \theta, y)
$$
Here, $\Delta$ denotes the set that constrains the perturbations to ensure they remain limited in size. The most prevalent choice is the $\ell_\infty$-ball, which limits the maximum change for any single pixel:
$$
\Delta = \left\{ \delta \mid \|\delta\|_\infty \leq \epsilon \right\}
$$
where $\epsilon$ represents the perturbation budget, i.e., the maximum allowable change in pixel intensity.

For targeted attacks, the goal is not to cause a mistake, but to force the model to predict a specific target label $y_{\text{target}}$. This is achieved by optimizing a contrastive objective that pushes the model's prediction away from the true label and towards the target label:
$$
\max_{\delta \in \Delta} \left( \mathcal{L}(x + \delta, \theta, y) - \mathcal{L}(x + \delta, \theta, y_{\text{target}}) \right)
$$
The resulting adversarial example is defined as $x^* = x + \delta$. While the loss function $\mathcal{L}$ is generic, in modern object detectors it is typically factorized into three main terms: classification, localization, and objectness. \\
\textbf{Classification Loss} ($\mathcal{L}_{\text{cls}}$): Measures the error in predicting the class of a detected object, for example, using cross-entropy.
$$
\mathcal{L}_{\text{cls}} = -\sum_{i \in \mathcal{O}} \sum_{c=1}^{C} y_i^{(c)} \log \hat{p}_i^{(c)}
$$
Here, $\mathcal{O}$ is the set of anchors matched to ground-truth objects, $C$ is the number of classes, $y_i^{(c)}$ is the binary ground-truth indicator, and $\hat{p}_i^{(c)}$ is the predicted probability that object $i$ belongs to class $c$. \\
\textbf{Localization Loss} ($\mathcal{L}_{\text{loc}}$): Measures the error between the predicted bounding box coordinates ($\hat{b}$) and the ground-truth box ($b$), for example, using the Smooth L1 loss.
$$
\mathcal{L}_{\text{loc}} = \sum_{i \in \mathcal{O}} \sum_{j \in \{x, y, w, h\}} \text{SmoothL1}(b_i^{(j)} - \hat{b}_i^{(j)}) \quad \text{where} \quad
\text{SmoothL1}(z) =
\begin{cases}
0.5 z^2 & \text{if } |z| < 1 \\
|z| - 0.5 & \text{otherwise}
\end{cases}
$$
where $j$ iterates over the bounding box parameters: center coordinates $(x, y)$, width $(w)$, and height $(h)$. \\
\textbf{Objectness Loss} ($\mathcal{L}_{\text{obj}}$): Measures the error in determining whether a given bounding box contains an object, for example, using binary cross-entropy.
$$
\mathcal{L}_{\text{obj}} = -\sum_{i=1}^{N} \left[ y_i^{\text{obj}} \log \hat{o}_i + (1 - y_i^{\text{obj}}) \log(1 - \hat{o}_i) \right]
$$
where $N$ is the total number of candidate boxes, $y_i^{\text{obj}} \in \{0,1\}$ indicates the presence of an object, and $\hat{o}_i$ is the predicted objectness confidence score.

\subsubsection{A taxonomy of attacks and models for object detection}

\begin{figure}[htbp]
    \centering
    \includegraphics[width=1\textwidth]{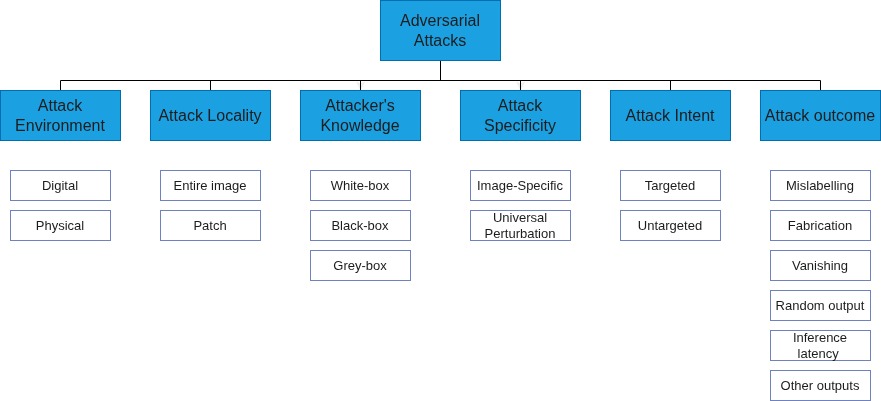}
    \caption{A taxonomy of adversarial attacks in object detection.}
    \label{fig:attack_taxonomy}
\end{figure}

Attacks and models can be classified based on their characteristics, context, and intent (see Figure \ref{fig:attack_taxonomy}). The outcomes of attacks, i.e., the types of impact that the adversarial perturbation has on the predictions of the target detector, are then used to classify the attacks in our survey. \\

\texttt{Attack environment}: \textbf{Digital}: Attacks that directly distort the input image of the target detector. This is by far the most common type of attack in the literature \citep{EBAD, DAG, OSFD}, applicable to any suitable existing dataset. \\
\textbf{Physical}: Attacks that involve modifying or adding real-world objects \citep{AdvTshirt} to the scene captured by a camera to fool the detector. They are usually an extension of digital attacks, especially for those using a patch that can be printed or projected \citep{InvCloak}. \\

\texttt{Attack locality}: \textbf{Entire Image}: The perturbations are not restricted to any part of the image. These attacks usually aim to remain "imperceptible" and are often constrained by a perturbation budget limit to control the alteration \citep{EBAD, DAG, OSFD}. \\ 
\textbf{Patch}: The effects of these attacks are restricted to a specific area. Some attacks allow multiple patches on one image or patches with different dimensions. These patches are generally not intended to be imperceptible, but some are, such as PRFA \citep{PRFA}, Imperceptible background patches \citep{ImpBackPatches} or RPAttack \citep{RPAttack}.  \\

\texttt{Attacker's knowledge}: \textbf{White-box attacks}: In a white-box attack scenario, the attacker has complete knowledge of the target model, including the network architecture, the parameter values, the training procedure and the training dataset. Therefore, specific attacks can be designed to maximize the effects on the target \citep{TOG}. \\
\textbf{Black-box attacks}: In a black-box attack scenario, the attacker does not have knowledge of the target model. They are limited to querying the model with images and observing the resulting predictions. To overcome this, many methods rely on transferability: some attacks utilize a known "surrogate" model to generate perturbations that target the unknown "victim" \citep{OSFD}. Others employ an ensemble of surrogate models to ensure the perturbation remains effective across different types of detectors, such as CAA \citep{CAA} or EBAD \citep{EBAD}. \\ 
\textbf{Grey-box attacks}: A grey-box attack scenario falls between white-box and black-box attacks. In this setting, the attacker only has some knowledge about the target model. For example, only the network architecture may be known, but not the training dataset or the parameter values. This additional incomplete information can enable more effective attacks than pure black-box methods \citep{EBAD}.  \\

\texttt{Specificity}: \textbf{Image-specific}: The attack applies an optimized perturbation which is different for each image. This is the most common paradigm among existing attacks \citep{DAG, EBAD, OSFD}.  \\
\textbf{Universal perturbation}: The same perturbation is applied to every image in the dataset. This process requires a training dataset to generate the perturbation, usually with a distribution close to the test dataset. Universal attacks are generally more generic, yet less effective than image-specific ones. Examples include U-DOS \citep{U-DOS} or PhantomSponges \citep{PhantomSponges}.  \\

\texttt{Intent}: \textbf{Targeted}: Usually associated with mislabeling attacks such as EBAD \citep{EBAD}. In a targeted attack, the attacker can choose which class will be predicted after applying the perturbation.  \\
\textbf{Untargeted}: The attacker does not specify a target class for the attack. All random output attacks are considered non-targeted \citep{OSFD}.  \\

\begin{figure}[t]
    \centering
    \includegraphics[width=\textwidth]{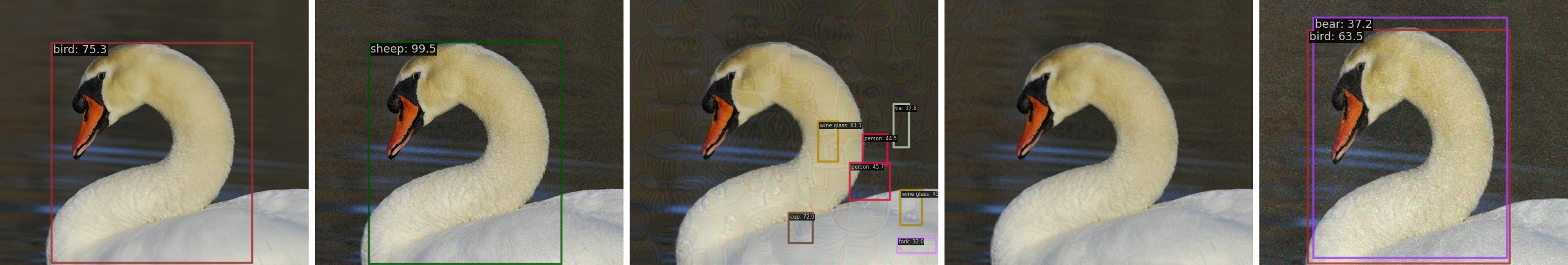}
    \caption{Inference results for different attack outcomes. From left to right: clean image, object mislabeling attack (EBAD \citep{EBAD}), random output attack (OSFD \citep{OSFD}), object vanishing attack, and object fabrication attack (PhantomSponges \citep{PhantomSponges}).}
    \label{fig:outcomesexamples}
\end{figure}

\texttt{Outcome}: \textbf{Object mislabeling}: A mislabeling attack perturbs the image so that the model predicts the wrong class for one or more objects present \citep{CAA, EBAD}. \\
\textbf{Object fabrication}: A fabrication attack is designed to make the model detect non-existent objects in the input image \citep{Daedalus}. Sometimes, it can be done with the specific intention of preserving the detection of existing objects, as with PhantomSponges \citep{PhantomSponges}.  \\
\textbf{Object vanishing}: A vanishing attack aims to erase one or multiple objects from the model predictions \citep{AdvART, ASC}. \\
\textbf{Random output}: A random attack alters the images so that the predictions of the model are different from the ground-truths. There is no specific intent for the result; some objects may vanish, be fabricated, and be mislabeled \citep{OSFD}.  \\
\textbf{Inference latency}: The purpose of these attacks is to increase the target model's inference time. This is more of an additional effect and is not a common type of attack. As an example, PhantomSponges causes inference latency in addition to having an object fabrication outcome \citep{PhantomSponges}.  \\
\textbf{Other outputs}: This category encompasses attacks whose results do not fit into the outcomes mentioned above. For example, ShiftAttack \citep{ShiftAtt} which only shifts the predicted bounding boxes, or attacks that can be used in several configurations for different purposes. Examples include TOG, which can be used as a random, vanishing, fabrication, or mislabeling attack with different constraints of locality and specificity \citep{TOG}, or ADC, which can be a mislabeling, vanishing, or fabrication attack \citep{ADC}. \\
An example of an inference result for each outcome is shown in Figure \ref{fig:outcomesexamples}. 

\subsubsection{Other definitions}

\textbf{Transferability}: The capacity of a perturbation initially crafted for one model to remain effective against another model or in different configurations and situations. Transferability can be assessed cross-model, cross-backbone, cross-dataset, cross-domain, etc. It is often evaluated in grey-box or black-box settings \citep{CAA, EBAD}. \\

\texttt{Attacked detector types}: \textbf{One-stage}: A one-stage detector directly predicts bounding boxes and class labels in a single pass, typically favoring speed.
\textbf{Anchor-based one-stage}: These use predefined anchor boxes from which final predictions are refined. Examples include early YOLO versions like YOLOv4 \citep{Yolov4}, SSD \citep{SSD}, and RetinaNet \citep{RetinaNet}.
\textbf{Anchor-free one-stage}: These directly predict object properties, like center points and dimensions, without predefined anchors, simplifying the pipeline. Examples include FCOS \citep{Fcos}, CenterNet \citep{CenterNet}, and recent YOLO versions (e.g., YOLOX \citep{YOLOX}). This also covers transformer-based detectors like DETR \citep{DETR} and Deformable-DETR \citep{DeformableDETR}, which directly predict objects. 

\textbf{Two-stage}: A two-stage detector first generates region proposals and then classifies them, usually prioritizing accuracy over speed.
\textbf{Anchor-based two-stage}: These primarily use a Region Proposal Network (RPN) to generate proposals based on predefined anchor boxes, refined in a second stage. Examples are Faster R-CNN \citep{faster-r-cnn}, Cascade R-CNN \citep{Cascader-cnn}, or Libra R-CNN \citep{Librar-cnn}.
\textbf{Anchor-free two-stage}: Less common than anchor-based for the two-stage paradigm, this approach involves generating initial region proposals without using predefined anchors and then refining the results. Examples in this category include Sparse R-CNN \citep{Sparser-cnn} and QueryDet \citep{Querydet}.

\subsection{A review of adversarial attacks}\label{sec:reviewATT}

Unlike attacks on image classification, object detection attacks are more complex due to the multi-objective nature of the task, involving both object localization and classification. These attacks can be broadly categorized based on their intended outcome.

\subsubsection{Object mislabeling}\label{sec:reviewATTmiss}

Inspired by classification attack methods, \emph{Dense Adversary Generation (DAG)} is one of the first attacks to target object detection using a gradient-based optimization process \citep{DAG}. Focusing on more specific object detection applications, \emph{ShapeShifter} \citep{ShapeShifter} can generate disturbed "stop" signs that are mis-predicted, highlighting vulnerabilities in autonomous driving vehicles. \emph{Universal Physical Camouflage Attacks} \citep{UPC} create patterns applied to people's clothes, thus deceiving the target detector towards the desired class. \emph{Pick-Object Attack} \citep{Pick-Object} allows for more precise disruptions by targeting only one object per image to change its class. Later attacks also use and alter the context of objects in images. Examples are \emph{Zero-Query Transfer Attacks (ZQA)} \citep{ZQA}, designed specifically for black-box applications, and \emph{Targeted context attack} \citep{TargConAtt} that focuses on altering the context information of objects by adding a context classification network which brings the original context information closer to the target's contextual information. A key limitation of many  methods is their poor transferability in a black-box setting, which has been improved in the two following methods. \emph{Context-Aware Transfer Attack (CAA)} \citep{CAA} also leverages contextual dependencies such as object co-occurrence, relative positions while using an ensemble of surrogate models to achieve high transferability. \emph{Ensemble-based Black-box Attacks on Dense Prediction (EBAD)} \citep{EBAD} is another ensemble-based attack that optimizes the contribution of each surrogate model for perturbation creation to maximize black-box success.

\subsubsection{Object vanishing}

Many attacks have been designed to make objects disappear. Several methods aim to make people vanish from detectors such as \emph{Invisible Cloak} \citep{InvCloak}, \emph{Adversarial T-shirt} \citep{AdvTshirt}, \emph{Patches to attack person detection} \citep{PatchesPerson} and \emph{FB Invisible cloak} \citep{FbInvCloak}. They involve patterns on clothing as well as printable patches. Other patch attacks have been developed for objects other than people. \emph{Contextual Adversarial Patches} \citep{ContAdvPatches} uses spatial context elements to fool the detector with a patch that does not overlap with objects, and \emph{Diffused Patch Attack (DPattack)} \citep{DPAttack}, a diffused asteroid-shaped or grid-shaped patch that only changes a small number of pixels to fool detectors. Recently, \emph{Transfer-based Self-Ensemble Attack (T-SEA)}  \citep{T-SEA} locates every target object from the images with a white-box detector and attaches adversarial patches to their center. \emph{Adversarial Art (AdvART)} \citep{AdvART} generates semantically coherent patches that attack object detectors. By enforcing semantic constraints, it outperforms methods based on Generative Adversarial Networks (GANs), making the patch more effective and realistic.

For attacks affecting the entire image, \emph{Evaporate Attack} \citep{Evaporate} combines genetic algorithm and particle swarm optimization to target both region-based and regression-based object detectors. Similarly, \emph{Contextual Adversarial Attack (CAP)} \citep{CAP} damages context information to achieve high transferability performance and \emph{Universal Dense Object Suppression (U-DOS)} generates universal perturbations while remaining highly effective in making predictions disappear \citep{U-DOS}. As for patch-based attacks, \emph{Refined Patch Attack (RPAttack)} \citep{RPAttack} only affects a few pixels of the object and \emph{Naturalistic Physical Adversarial Patch} \citep{NaturalisticPatches} generates natural-looking adversarial patches that remain visually inconspicuous using GAN-based optimization. In more specific applications, \emph{Adversarial Texture} \citep{AdvTexture} applies patterns to people's clothes to fool person detectors even under different viewing angles and deformations. \emph{Patch Attack Against Unmanned Aerial
Vehicles (UAV)} \citep{PatchesUAV} uses patches placed on cars to hide them from detectors. The most recent vanishing attack that does not involve a patch is \emph{Adversarial Semantic Contours (ASC)} \citep{ASC}. Instead of modifying the entire image, ASC only perturbs a small number of pixels, around the object's contour, to fool the detector and make the object disappear.

\subsubsection{Object fabrication}

Fabrication attacks often target the non-maximum suppression (NMS) step, which eliminates redundant bounding boxes around the same object, in order to create non-existent objects. For example, \emph{Daedalus} \citep{Daedalus} attacks the NMS by compressing the dimensions of the bounding boxes such that detections do not overlap and creates extremely dense false positives. \emph{PhantomSponges} \citep{PhantomSponges} generates a universal adversarial perturbation that also targets NMS. This perturbation causes inference latency by adding non-existent objects that overload the NMS. For patch-based attacks, \emph{Patch-based False Positive creation} \citep{PatchFP} generates a series of patches, each designed to cause the detection of a non-existent object of a different class. 

\subsubsection{Random output}

Some attacks do not have a specific outcome objective, such as \emph{Robust Adversarial Perturbation (R-AP)} \citep{R-AP} that aims to disturb the predictions of the Region Proposal Network. Inspired by the mislabeling attack DAG (see Section \ref{sec:reviewATTmiss}), \emph{Unified and Efficient Adversary (UEA)} \citep{UEA} optimizes a total loss function which is a combination of a GAN loss and the DAG attack loss. The GAN framework generates perturbations that are effective and visually similar to the original images. For patch-based attacks, \emph{DPatch} \citep{DPatch} produces location independent patches, while \emph{Imperceptible background patches} \citep{ImpBackPatches} applies several imperceptible patches in the background of images. \emph{Parallel Rectangle Flip Attack} \citep{PRFA} also uses imperceptible patches around objects placed at critical points, avoiding suboptimal detection near the attacked region. \emph{Relevance Attack on Detectors (RAD)} \citep{RAD} aims to achieve high transferability by attacking the relevance map, a common property across interpreters in detectors. \emph{Local Perturbations with Adaptively Global Attacks (LGP)} \citep{LGP} is a detector-agnostic method that works across different models and datasets. LGP focuses perturbations on object regions, avoiding unnecessary changes in background areas. A recent method that exhibits high transfer performance is \emph{Object-Aware Significant Feature Distortion (OSFD)} \citep{OSFD}. It combines a loss function and a designed data augmentation method that suppresses significant features and amplifies the vicinal features to create false detections that transfer well to other detectors.

\subsubsection{Other outputs}

Some attacks can be used in several configurations with different outcomes. \emph{Seeing isn’t Believing} \citep{SeeingisntBelieving} is composed of two attacks using feature interference reinforcement (FIR) and enhanced realistic constraints generation (ERG) for the vanishing one and nested-AE for the fabrication one. FIR deals with interfering with the internal features of the model to increase robustness, while ERG focuses on making the generated adversarial example more realistic and adaptable to real-world conditions. This approach helps the attack remain effective across different distances. \emph{Targeted Adversarial Objectness Gradient Attacks (TOG)} \citep{TOG} is a suite of adversarial attacks that exploits gradients of the detector loss function to generate different perturbations with different outcomes: random output, object vanishing, object fabrication, and mislabeling.

\emph{Adversarial Attacks That Evade Context Consistency Checks (ADC)} is a method that generates adversarial images to both deceive object detectors and bypass context consistency checks \citep{ADC}. ADC can be deployed as a mislabeling, vanishing, or fabrication attack. Addressing the specific challenge of attacking modern transformer-based detectors, the \emph{Attention-Focused Offensive Gradient (AFOG)} attack \citep{AFOG} uses a learnable attention mechanism to dynamically focus perturbations on vulnerable image regions for random output generation.  AFOG also includes specialized modes for vanishing and fabrication.\\
\emph{ShiftAttack} \citep{ShiftAtt} has a unique objective where it targets post-processing techniques, such as Non-Maximum Suppression, to generate the adversarial image. It exploits predictions with low intersection over Union in the detectors and boosts their confidences so that they replace the ones with higher IoU, leading to mislocalization and missed detections.

\subsection{A review of adversarial defense methods}\label{sec:reviewDEF}

Countermeasures against adversarial attacks on object detectors can be categorized by the stage at which they intervene: during model training, by modifying the model's architecture or inputs, or by adding external detection modules. These approaches are sometimes adapted from fundamental defense strategies from image classification to the more complex task of object detection.

\subsubsection{Adversarial training and training modification}

Adversarial training remains the most effective and researched defense strategy. This approach involves augmenting the training set with adversarial examples, thereby making it more adversarially robust.

One of the initial efforts to adapt this technique for object detection proposed generating adversarial examples through the lenses of a multitask learning problem \citep{AdvtrainZhang}. Adversarial training was further improved through the following contributions. \emph{Class-Aware Robust Adversarial Training (CWAT)} was introduced to solve the class imbalance problem \citep{AdvtrainChen}, which was not taken into account by naive adversarial training,. This method implements a class-wise regularization term to account for class imbalances during training. Other methods seek to refine the learning process itself. \emph{RobustDet} introduces adversarially-aware convolutions to disentangle feature learning between clean and adversarial images combined with an Adversarial Image Discriminator (AID) \citep{RobustDet}. However, later research suggests the AID component is primarily effective against strong perturbations and less so for weaker ones \citep{AdvRecent}. A recent and highly effective approach leverages adversarially pre-trained backbones; for instance, it has been demonstrated that initializing an object detector with a backbone pre-trained on adversarial examples, rather than clean ones, significantly boosts adversarial robustness without altering the detector’s architecture \citep{AdvRecent}.

There also exist instances where adversarial training is used to defend against adversarial patches. For example, \emph{Meta Adversarial Training (MAT)} \citep{AdvMeta} combines adversarial training with a meta-learning framework for universal patches, thereby increasing robustness against universal patch attacks.

There also exist approaches that modify the training process. For example, it has been observed that detectors can be fooled by patches placed outside an object's bounding box, which exploits the model's reliance on spatial context \citep{ContAdvPatches}. To mitigate this, the authors proposed limiting the use of spatial context during training, making the detector less dependent on potentially exploitable contextual cues and more robust to such attacks.

\subsubsection{Model modification and input transformation}

Beyond data augmentation and training process, modifying the model’s intrinsic architecture or adding components that will transform the inputs allows for the creation of robust models that are trained on standard data.

Architecture transformation can make models intrinsically more robust at the cost of often making the model more complex. One such modification involves integrating \emph{Gabor convolutional layers} into the backbone of object detectors \citep{AdvtrainGabor}. These filters extract robust spatial features, and by training them as standard network parameters in the first layer, this method enhances resilience to adversarial attacks without degrading performance on benign inputs.

Input transformation acts as a preprocessing step to filter an image before it is fed to the detector. This approach can be model-agnostic. Simple methods like JPEG compression can incidentally degrade adversarial perturbations; however, they are not a complete defense by themselves and must be combined with other methods. More sophisticated techniques employ dedicated models for purification. One such defense utilizes an autoencoder framework coupled with a critic network during training \citep{filter}. In this setup, the critic guides the encoder to map both benign and adversarial inputs to a common, robust feature distribution. The decoder then reconstructs a "clean" version of the image from this robust representation, which is then passed to the object detector.

\subsubsection{Adversarial detection}

Rather than purifying an image, some defenses focus on detecting whether an input is adversarial or not. These methods act as a gatekeeper, flagging and rejecting suspicious inputs.

The \emph{SCEME} system was proposed as a method to detect adversarial examples by identifying context inconsistencies \citep{ContextInconsistency}. This approach identifies abnormal statistical relationships between objects and their context that typically arise from adversarial perturbations. \emph{SCEME} learns these context consistency rules during training using a set of class-specific auto-encoders. At inference time, a context profile is extracted for each detected region and fed to the corresponding auto-encoder.

Most adversarial detection methods focus on adversarial patch detection. Early research in this area focused on identifying patches based on their underlying feature representations \citep{AdvPatchFeatEnergy}. Another strategy is to explicitly segment the patch and remove it as employed in \emph{Segment and Complete (SAC)} \citep{SAC} which trains a patch segmenter with self-adversarial training for pixel-level localization, after which a shape completion algorithm inpaints the removed area. However, its reliance on training with noisy patches limits its effectiveness against more sophisticated, natural-looking patches.

More recent, model-agnostic methods have shown greater promise against realistic patches. \emph{Jedi} \citep{JEDI} leverages entropy analysis to identify high-entropy regions likely to contain a patch. It refines this localization using an autoencoder to complete these regions. Building on this, \emph{Patch-Agnostic Defense (PAD)} \citep{PAD} operates without any training by leveraging two universal properties of patches: semantic independence from the scene and spatial heterogeneity compared to their immediate surroundings.

\subsubsection{Certified defenses}

While the previous methods show empirical success, certified defenses provide a formal, mathematical guarantee of robustness against a specific class of attacks within a defined perturbation budget ($L_p$-norm ball). The first model-agnostic and training-free certified defense for object detection against $L_2$-bounded attacks was established by reducing the detection task to a regression problem and applying median smoothing \citep{CertDet}. 

For the specific threat of patch attacks, \emph{DetectorGuard} \citep{detectorguard} employs an "objectness explaining strategy", using a robustly trained image classifier to scan every location in an image and predict the probability of an object's presence, thereby providing a provable security guarantee against patch hiding attacks. A subsequent work, \emph{ObjectSeeker} \citep{objectseeker}, neutralizes patches by masking them entirely without prior knowledge of their appearance or location, and includes a certification procedure to formally guarantee robustness.

\subsection{Evaluating attack impact and perceptibility}
\subsubsection{Evaluation metrics}\label{sec:surveyperfmetrics}

In order to measure the impact of an attack, it is necessary to use appropriate metrics. Unlike classification, object detection involves localizing and classifying multiple objects, leading to a more complex performance evaluation. Furthermore, the diverse objectives of different adversarial attacks often require tailored metrics. This section presents the most common metrics among the attack and defense methods listed in Sections \ref{sec:reviewATT} and \ref{sec:reviewDEF} and highlights their applicability and limitations. \\

\textbf{mean Average Precision (mAP) \& Average Precision (AP)}: The AP is calculated by averaging the precision-recall curves for an object class. \textit{Precision} quantifies the ratio of true positives out of all detected objects (True positives and False positives), while \textit{Recall} measures the fraction of true positives out of all actual objects in the image(true positives and false negatives). The mAP is simply the mean of the APs for each object class. When computing this metric, a prediction is considered as valid for a ground truth if their Intersection over Union (IoU) is greater than a chosen threshold. Generally, to calculate the map, this threshold is set at 0.5 \citep{OSFD, LGP, CAP}. For adversarial attacks, a significant drop in mAP indicates a successful attack, as it reflects a degradation in the detector's localization and classification performance. mAP is a comprehensive metric for detector performance, however it does not precisely isolate the specific objective of certain adversarial attacks. \\

\textbf{Success rate/Fooling ratio/Attack Success rate (ASR)}: The ratio of successfully attacked images. The "success" of an attack is a subjective matter, often adapted to the type of attack being studied which makes comparison across papers challenging and hinders benchmarking attempts. In practice, the success rate is used more often for vanishing and mislabeling attacks. For example, in EBAD the Fooling rate is the ratio of images in which the victim object was successfully misclassified at IoU greater than 0.3 to the target label \citep{EBAD} whereas in ASC, a prediction is considered as correct if it has an IoU above 0.5 and a confidence score above a threshold depending on the detector used \citep{ASC}. In addition, the Attack Success rate from ShiftAttack is very specific. As it measures the proportion of objects not assigned by any true positive prediction box, but also with the existence of a false positive box of the same label having a valid IoU \citep{ShiftAtt}. \\

\textbf{False Positive rate (FP rate)}: The ratio of false positive detections to
the total number of predictions. This metric is used mainly by object fabrication attacks \citep{Daedalus, PatchFP}, where creating false detections is the main objective.

\subsubsection{Evaluation of the magnitude of the perturbations}\label{sec:surveyperturcost}

Attacks intended to remain imperceptible to evade detection need to contain the magnitude of the perturbations they generate. Generally, a threshold is introduced during the optimization of the attack in order to limit the distance between the original image and the perturbed image according to a certain chosen norm. It is also possible to evaluate this distance after the creation of adversarial images using other norms. Here are defined the distances and norms most commonly used in the literature. \\

\textbf{$L_p$ Norms}: $L_p$ norms are the most commonly used to calculate the distances between the original images and the adversarial images, and often integrated into the optimization process to constrain the perturbation budget of the attack. Among them we have:
\begin{itemize}
    \item \textbf{$L_0$ norm:} Measures the number of perturbed pixels.
    $ \| \delta \|_0 = \sum_{i=1}^{N} \mathbb{I}(\delta_i \neq 0) $
    where $\delta$ is the perturbation vector and $N$ is the total number of pixels.
    \item \textbf{$L_1$ norm:} Calculates the sum of the absolute differences between corresponding pixel values.
    $ \| \delta \|_1 = \sum_{i=1}^{N} | \delta_i | $

    \item \textbf{$L_2$ norm (Euclidean distance):} Represents the square root of the sum of squared differences between pixel values. It reflects the average intensity of the perturbation.
    $ \| \delta \|_2 = \sqrt{\sum_{i=1}^{N} \delta_i^2} $

    \item \textbf{$L_\infty$ norm:} Quantifies the maximum absolute difference between any single pixel's value in the original and perturbed images.
    $ \| \delta \|_\infty = \max_{i} | \delta_i | $
\end{itemize}

The attacks are mostly optimized for $L_\infty$ or sometimes $L_2$ \citep{EBAD, CAA, PhantomSponges}. \\

\textbf{Peak-to-noise ratio (PSNR)}: A metric that quantifies the amount of perturbation or noise introduced into an image. A low PSNR corresponds to a highly distorted image, and vice versa. When it is used, the objective is, therefore, to maintain it above a chosen threshold \citep{CAP, ImpBackPatches, R-AP}. \\
It is computed according to the following formula:
\begin{equation}
    \text{PSNR}=10\log_{10}\frac{\text{MAX}^2}{\text{MSE}}
\end{equation}
where MAX denotes the maximum possible pixel intensity value, and MSE represents the mean squared error between the original and perturbed images. \\

\textbf{Structural Similarity Index Measure (SSIM)}: SSIM is a perception-based measure to estimate similarity between two images. It considers structural dependency, the idea that spatially-close pixels have strong inter-dependencies, and perceptual phenomena, such as luminance and contrast, rather than pixel-wise distances. In the approaches listed in Section \ref{sec:reviewATT}, SSIM is used after the computing of the adversarial images to estimate the magnitude of the perturbation in a way that more closely reflects human perception \citep{AdvART, Pick-Object, ShiftAtt}. LGP attack also uses Information-Weighted Structural Similarity Index (IW-SSIM), a version of the Structural Similarity Index (SSIM) incorporating an information-theoretic weighting scheme \citep{LGP}.

\subsection{Are there fair comparisons between attacks ?}\label{sec:criticsurvey}

Comparing different attack methods presented in the previous Section \ref{sec:reviewATT} is almost impossible. This is due to the lack of a shared benchmark and test method, including common metrics, datasets, and detectors, which reflect the impacts of all attack types rightfully. Table \ref{tab:subsetallATT} presents the most recent and advanced attacks, detailing their outcomes, specificities, locality, perturbation cost metrics, and test configurations reported in their original articles. Attacks selected for our unified benchmark are explicitly marked with a dagger ($\dagger$). Table \ref{tab:AllATT} in appendix lists this information for all attacks presented in Section \ref{sec:reviewATT}. 

\begin{sidewaystable}
\caption{Recent adversarial attacks for object detection: Heterogeneity in experimental settings and applicability hinders a fair comparison. * denotes if the code is publicly available. $\dagger$ denotes attacks that we used in our benchmark. Detector families used: Y: YOLO, F: Faster R-CNN, S: SSD, D: DETR, V: VFNet, C: Cascade R-CNN, M: Mask R-CNN, A: ATSS/Anchor-free, R: RetinaNet, G: Grid R-CNN, L: Libra/FoveaBox/FreeAnchor, T: Transformer-based (e.g., Swin, DAB-DETR, Def-DETR). See Table \ref{tab:AllATT} in Appendix \ref{app:survey_data} for a more exhaustive list}
\label{tab:subsetallATT}
\centering
\renewcommand{\arraystretch}{1.7}
\begin{footnotesize}
\begin{tabular}{ccccccc}
\toprule
NAME & OUTCOME & LOCALITY & \makecell{PERTURBATION \\ COST} & METRICS & DETECTOR FAMILY & DATASET USED \\
\midrule

AdvART \citep{AdvART} & Vanishing & Patch & SSIM & \makecell{mAP, \\ Success rate} & \textbf{Y} & \makecell{INRIA, MPII} \\

LGP \citep{LGP}* & Random & Entire image & \makecell{FID, IW-SSIM, \\ PSNR-B} & \makecell{mAP, Number \\of attacked targets per \\ image, Success rate} & \textbf{F, T, V, C, A} & \makecell{COCO, \\DOTA-v1.0} \\

OSFD \citep{OSFD}* $\dagger$& Random & Entire image & $L_\infty$ & mAP & \textbf{F, Y, V, M, D, A} & VOC2012 \\

\makecell{Patch-based FP \\ creation \citep{PatchFP}} & Fabrication & Patch & - & \makecell{mAP, \\ Average instances created, \\Average Precision Decrease, \\ Average Score Created, \\ FP Rate Increase} & \textbf{Y} & \makecell{DOTA-v1.0,\\RSOD, NWPU \\ VHR-10} \\

ShiftAttack \citep{ShiftAtt} & Other & Entire image & $L_2$, SSIM & Success rate & \textbf{Y, A, L, R, T} & \makecell{COCO, VOC2007, \\ VOC2012} \\

\makecell{Targeted context \\ attack \citep{TargConAtt}} & Mislabeling & Entire image & $L_\infty$ & Success rate & \textbf{F, Y, L, T} & \makecell{COCO, VOC2007} \\

ASC \citep{ASC} & Vanishing & Entire image & $L_0$ & Success rate & \textbf{S, Y, F, M, T} & \makecell{COCO, \\ Cityscapes, \\ BDD100K} \\

\makecell{CAA \citep{CAA}* $\dagger$} & Mislabeling & Entire image & $L_\infty$ & Success rate &  \textbf{F, Y, R, L, D, T} &  \makecell{COCO, VOC2007}\\

EBAD \citep{EBAD}* $\dagger$& Mislabeling & Entire image & $L_\infty$ & Success rate & \textbf{F, Y, A, S, G, R, L, T} & \makecell{COCO, VOC2007} \\

\makecell{Patches against UAV \\ \citep{PatchesUAV}*} & Vanishing & Patch & - & \makecell{mAP, \\ Success rate} & \textbf{Y} & \makecell{VisDrone-2019} \\

\makecell{Phantom \\ Sponges \citep{PhantomSponges}* $\dagger$} & \makecell{Fabrication, \\ Inference latency} & Entire image & $L_2$ & \makecell{Number of created objects, \\ time, recall} & \textbf{Y} & \makecell{BDD100K, MTSD, \\ LISA, VOC} \\

T-SEA \citep{T-SEA}* & Vanishing & Patch & - & \makecell{AP \\ (for person)} & \textbf{F, Y, S} & \makecell{INRIA, \\ COCO-person, \\ CCTV-person} \\

\botrule
\end{tabular}
\end{footnotesize}
\end{sidewaystable}

Regarding the \textbf{detectors} employed, there is no doubt that the YOLOv3 \citep{yolov3} and Faster R-CNN \citep{faster-r-cnn} detectors stand out as the most widely used, representing both one-stage and two-stage detectors. It is important to note that some attacks can only be generated by a single type of detector, such as PhantomSponges, which can only be generated with a YOLO \citep{PhantomSponges} or R-AP \citep{R-AP} that targets the Region Proposal Network, a component that is mostly present in two-stage detectors and absent from YOLOv3. However, even in these cases, it is always possible to test other detectors in a black-box configuration. 

For \textbf{datasets}, the vast majority of attacks use widely accepted and accessible datasets such as COCO \citep{COCO}, VOC2007 or VOC2012 \citep{VOC2007}. COCO-person, a subset of COCO dedicated to person detection, is mostly used by patch-based attacks \citep{T-SEA}. Nevertheless, attacks do not all use the datasets in the same way. E.g., OSFD takes a sample of only 2000 images from VOC2012 for its tests \citep{OSFD}. Attacks designed for specialized applications like aerial surveillance rely on domain-specific datasets like DOTA \citep{DOTA} as standard benchmarks are inappropriate \citep{PatchesUAV}.

In terms of perceptibility evaluation, the Euclidean norms $L_\infty$ and $L_2$ are prevalent. However, their application is not always consistent, leading to potential misinterpretations. A key ambiguity lies in the $\epsilon$ parameter. In EBAD \citep{EBAD}, $\epsilon$ denotes the strict $L_\infty$ bound of the final adversarial example. In contrast, CAA \citep{CAA} treats $\epsilon$ merely as a step-wise constraint during optimization, not a total budget. This inconsistency extends to OSFD which, when parameterized with $\epsilon=5$ shows a resulting perturbation magnitude of $L_\infty = 7$  \citep{OSFD}. As these examples show, the parameters are not used in the same way at all. Moreover, $L_\infty$'s relevance as a model for the human eye's sensitivity to perturbations can be largely disputed, and it would be preferable to turn to more perceptual distances \citep{LPIPS}, as can be seen in Section \ref{sec:benchmark}.

Finally, to evaluate performance, the mAP is an excellent metric that represents well the effectiveness of a detector. It is widely used in the literature \citep{OSFD,AdvART,T-SEA} but it is sometimes not suitable for representing the specific effect of an attack. Some articles therefore use a Success rate (ASR) which, as mentioned in Section \ref{sec:surveyperfmetrics}, is different for each objective and, therefore, makes direct comparison between articles impossible \citep{ShiftAtt, ASC, EBAD}. Thus, it is necessary to use more standardized metrics, which better describe how the detector is affected by the tested attack.

This lack of a shared benchmark (i.e., same datasets and relevant metrics) is hindering progress in the field as fair comparisons between attacks are not possible.

\section{A unified benchmark for adversarial attacks for object detection}
\label{sec:benchmark}

\subsection{Proposed unified benchmark}

To address the issues established in Section \ref{sec:criticsurvey}, this section moves beyond a simple survey to propose a unified benchmark. Our goal is to evaluate attacks under identical and controlled conditions. This framework ensures a fair comparison of white-box effectiveness and black-box transferability by standardizing the detectors, the datasets, and the attack settings. Crucially, it also introduces a common set of metrics to evaluate both attack success and the perceptual cost of the perturbations, allowing for a fair side-by-side analysis.

\subsubsection{Choice of metrics}

It is imperative to provide common metrics that fairly reflect the impacts of each attack, regardless of their outcomes or characteristics. Object detection involves two tasks: localization and classification. Adversarial attacks tend to disrupt either one of these tasks (localization with vanishing attacks, classification with mislabeling attacks) or both (e.g., random and fabrication attacks) (cf. Figure \ref{fig:outcomesexamples}). In addition to the standard metric mAP, which is used to estimate the general impact of an attack on object detection performance, we introduce two new complementary metrics reflecting the impact on localization and classification separately: $\text{AP}_{\text{loc}}$ and CSR defined below. These metrics draw inspiration from error analysis research for object detection \citep{TIDE, diagnosing} but have never been applied to adversarial attacks. \\

$\textbf{AP}_{\textbf{loc}}$: This metric is calculated in the same way as the mAP, except that all object classes are fused as a single class. Therefore, it reflects the localization capacity of the detector, i.e., its ability to detect the presence of an object. Consequently, this metric is particularly sensitive to vanishing attacks and fabrication attacks. \\


\textbf{Classification Success Ratio (CSR)}: This metric evaluates the detector’s classification capability independent of localization failures. Mathematically, this metric is equivalent to the micro-averaged recall of the detector. It measures the ratio of ground-truth objects detected with the correct class to the total number of ground-truth objects. For each ground-truth object, we select the highest-confidence detection that satisfies the localization constraint (IoU $>$ threshold) and verify that the detection class matches the ground-truth class. The CSR can measure misclassification errors not captured by the $\text{AP}_{\text{loc}}$ where the object is found but labeled wrong. This makes CSR suitable for quantifying the impact of mislabeling attacks. It is computed as:
\begin{equation}
    \text{CSR} = \frac{\sum_{c=1}^{C} TP_c}{N_{GT}}
\end{equation}
where $C$ is the number of classes, $TP_c$ is the number of correctly classified and localized True Positives for class $c$, and $N_{GT}$ is the total number of ground truth instances across the entire dataset.

Even if the norms $L_\infty$ and $L_2$ are very widespread, they do not necessarily represent how the human eye will perceive the perturbation applied to the image. Thus, we propose using perceptual metrics, which better model the perceived difference between the original and attacked images, such as SSIM mentioned in Section \ref{sec:surveyperturcost} or LPIPS \citep{LPIPS}. These metrics are used in the same manner, between the original image and the final adversarial image, to achieve a fair comparison. \\

\textbf{Learned Perceptual Image Patch Similarity (LPIPS)} Evaluates the perceptual similarity between two images, better aligning with how humans perceive the images. The metric is based on the findings in \citep{LPIPS} that internal activations of deep networks (e.g., AlexNet, VGG) trained on classification tasks strongly correlate with human perception. It operates by computing the $L_2$ distance between the deep feature embeddings of two images.

\subsubsection{Choice of detectors}\label{sec:3expesetup}

To test our benchmark, we use the MMDetection framework, which allows us to easily use a large number of object detectors \citep{mmdetection}. We select a \textbf{set of detectors} to represent most detector types: YOLOv3 \citep{yolov3} and Faster R-CNN \citep{faster-r-cnn} as they are by far the most used detectors of their categories. YOLOX \citep{YOLOX} serves as a modern YOLO version. FCOS is an anchor-free detector. DETR \citep{DETR} is a standard transformer-based detector. DINO \citep{dino} represents modern, fully transformer-based architectures. Mask R-CNN \citep{maskrcnn} is included as another example of a model with a transformer backbone. The \textbf{backbones} used are Darknet-53 for YOLOv3 and ResNet-50 \citep{resnet} for Faster R-CNN, FCOS and DETR, CSPNet for YOLOX and Swin transformer \citep{swin} for Mask R-CNN and DINO.

\subsubsection{Choice of datasets}

All detectors are trained on COCO \citep{COCO} and tested on the VOC2007 test set \citep{VOC2007}. This is a standard evaluation setup for most attacks, as COCO-trained models are widely available (e.g., in MMDetection), attacks are often evaluated on VOC and the classes in VOC are a compatible subset of the COCO classes. 

\subsubsection{Choice of attacks}\label{sec:choice}

To ensure a fair and rigorous comparison, we focus on digital, non-patch-based attacks. Indeed, first, physical attacks are excluded as they are difficult to reproduce consistently. Their success depends on real-world variables (e.g., printability, lighting, camera angles) that cannot be standardized within a digital benchmark. Second, we exclude patch-based attacks. These attacks are perceptible by design and operate on a "cost" model (e.g., patch size, shape, or semantic realism) that is completely different from the perturbation measures used by imperceptible attacks. By focusing on digital, non-patch-based attacks, we can use a common set of metrics (like $L_\infty$, $L_2$, and LPIPS) to meaningfully evaluate the attack impact and perceptual cost.  \\
Within this scope, we select the most effective and recent attacks with available code. The only category not represented is the object vanishing one because no recent attack has available code. This gives us the selection: OSFD \citep{OSFD} which reported state-of-the-art attack efficiency for random output, EBAD \citep{EBAD}, an established mislabeling attack \citep{BestSurvey} and PhantomSponges \citep{PhantomSponges} for object fabrication. We also include CAA \citep{CAA} to study its unique focus of leveraging contextual information.

We initially considered an earlier object detection attack, TOG \citep{TOG} due to its importance in the literature, but decided against including it. The primary reason is that early attacks like TOG exhibit poor transferability even when the target model belongs to the same model family as the source model or with different implementations of the exact same model (cf. Table \ref{tab:tog_results}). This is largely due to a different research focus at their time where achieving attack success on a white-box setting was the primary objective and the transferability of the attack only analyzed qualitatively in the original paper. 

\begin{table}[ht]
\centering
\caption{Performance (mAP in \%) metric of YOLOv3 attacked by each TOG variant. ML and LL refer to Most Likely (2nd most probable) and Least Likely class selection strategies.}
\label{tab:tog_results}
\begingroup
\footnotesize
\renewcommand{\arraystretch}{1.3}
\begin{tabular}{@{}l|cccccc@{}}
\toprule
\textbf{Detectors} & \textbf{Benign} & \textbf{Untargeted} & \textbf{Fabrication} & \textbf{Vanishing} & \makecell{\textbf{Mislab.} \\ \textbf{(ML)}} & \makecell{\textbf{Mislab.} \\ \textbf{(LL)}} \\
\midrule
YOLOv3 (MMDet) & 75.84 & 59.47 & 68.15 & 66.87 & 70.24 & 69.68 \\
YOLOv3 (TOG)   & 78.72 & 1.88  & 37.09 & 17.22 & 20.53 & 16.44 \\
\botrule
\end{tabular}
\endgroup
\end{table}

We also evaluated the recent AFOG attack \citep{AFOG}, which initially seemed promising, as it is one of the few attacks generated directly on a transformer-based detector. However, similar to older attacks like TOG, the original paper does not focus on the transferability of the generated perturbations. The attack procedure involves resizing the input image during preprocessing. As shown in Table \ref{tab:afog_results}, while the attack is highly effective on the resized image (13.23\% mAP), its effectiveness collapses when the final adversarial image is resized back to the original input dimensions (66.56\% mAP). This extreme sensitivity indicates that the generated perturbation is brittle and unlikely to transfer well under different conditions, making it unsuitable for a benchmark focused on practical, transferable robustness. We ultimately excluded AFOG from our main benchmark due to its lack of robustness under minor transformations as it is unsuitable for a benchmark focused on transferability.

\begin{table}[h!]
\centering
\caption{The mAP (\%) metric of the AFOG attack (with $L_\infty=30$) against DETR in white-box. Attack efficiency significantly degrades when the attacked image of the VOC2007 test set are resized back to their original dimensions.}
\label{tab:afog_results}
\small
\begin{tabular}{@{}l|c@{}}
\toprule
\textbf{Attack Configuration} & \textbf{mAP (\%) on DETR} \\
\midrule
AFOG DETR ($L_\infty=30$) - Attacked Image Size & 13.23  \\ 
AFOG DETR ($L_\infty=30$) - Resized to Original Size &  66.56 \\ 
\bottomrule
\end{tabular}
\end{table}

\subsubsection{Implementation details}

For all attacks, we test the default parameter configuration of the original articles. For EBAD and CAA, where the main parameters $\epsilon_{EBAD}$ and $\epsilon_{CAA}$ influence the $L_\infty$ size of the perturbations, we also run tests for other values. As mentioned in Section \ref{sec:criticsurvey}, $\epsilon_{EBAD}$ represents the distance between the original image and the adversarial image, while $\epsilon_{CAA}$ represents the norm of the perturbation applied to each iteration of the attack \citep{EBAD,CAA}. We refer to both parameters simply as $\epsilon$ in the remainder of the paper to denote the attack strength. In addition, the parameter $k$ of OSFD monitors the amplification of object features \citep{OSFD}, and parameters $\epsilon_{PS}$, $\lambda_1$ and $\lambda_2$ of PhantomSponges influence the norm $L_2$ of perturbations, the number of objects transmitted to the NMS, and the minimization of IoUs \citep{PhantomSponges}. These parameters are set to their default values in our experiments: $k=3$, $\epsilon_{PS} = 70$, $\lambda_1=1$, $\lambda_2=10$. The IoU threshold is set at 0.5 to compute mAP, $\text{AP}_{\text{loc}}$ and CSR. For the attack time generation, we deactivated all logging, saving and evaluating steps to ensure fair comparison.

\subsubsection{Scenarios}

We perform tests in white-box and black-box scenarios for YOLOv3, Faster R-CNN and Mask R-CNN. These models are commonly used in the literature and are compatible with the codebases of most selected attacks. For black-box testing, DETR, FCOS, YOLOX, and DINO are used only as target detectors. As PhantomSponges can only be generated on a YOLO detector and has to be trained on a set of images different from the test images, this attack is generated with a YOLOv3 using the VOC2007 trainval set \citep{PhantomSponges}.

\begin{figure}[htbp]
    \centering
    \includegraphics[width=1\textwidth]{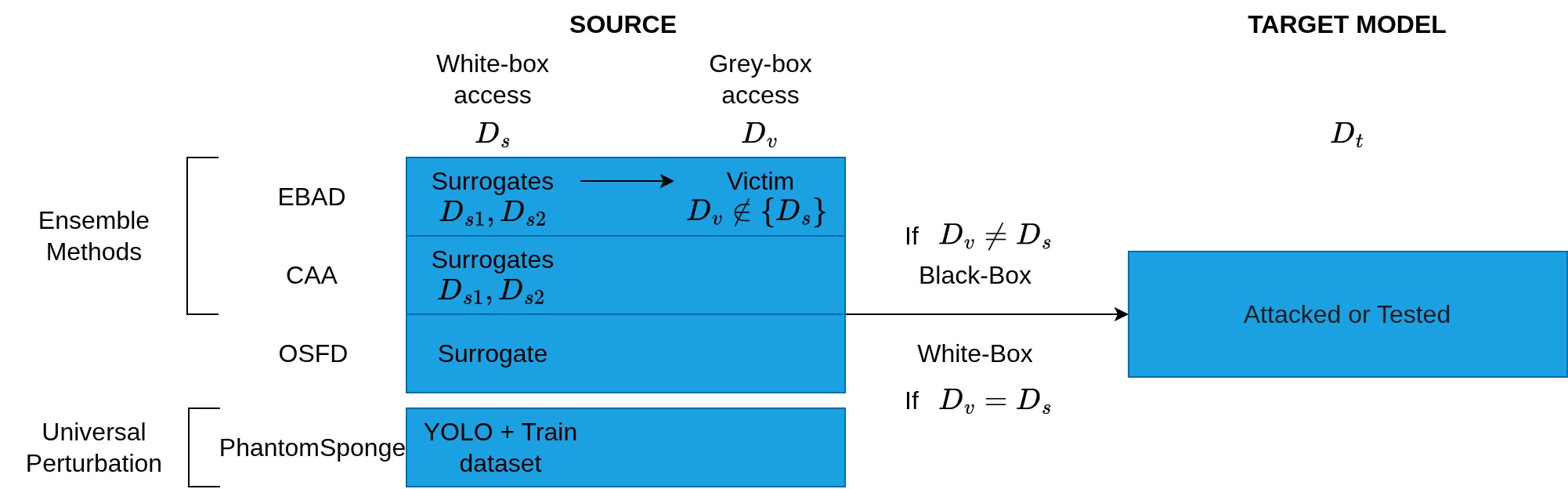}
    \caption{Threat models for the selected attacks. This diagram illustrates the source models and knowledge access required for generating each attack. $D_s$ denotes the \textbf{surrogate models} used for attack generation, while $D_v$ represents the \textbf{victim model} queried during the optimization of grey-box attacks (e.g., EBAD). $D_t$ refers to the \textbf{target model} used for final evaluation.}
    \label{fig:threat_model}
\end{figure}

For CAA, which is ensemble-based, we consider Faster R-CNN and YOLOv3 as surrogate models, so it uses both these models in a white-box setting. A specificity of the ensemble-based method EBAD is that it is query-based. This means that EBAD uses attacks created on the surrogates and then iteratively queries a "victim model" until the refined attack successfully fools the victim model. To refine the attack, the algorithm uses the inference loss of the victim model. Because it requires access to this loss value, EBAD operates in a grey-box setting, where the attacker has partial information about the victim model. To avoid any kind of confusion in our experiments, for attack generation, we use a white-box setting: The victim model queried by EBAD is one of the surrogate models (Faster R-CNN or YOLOv3). In addition, we test the transferability of this generated attack in a black-box setting. Figure \ref{fig:threat_model} summarizes these interactions between source, victim, and target models for each threat model.

\subsection{Results and analysis}

\subsubsection{Quantitative analysis of adversarial perturbation imperceptibility}\label{sec:benchdist}

\begin{figure}[htbp]
    \centering
    \includegraphics[width=0.8\textwidth]{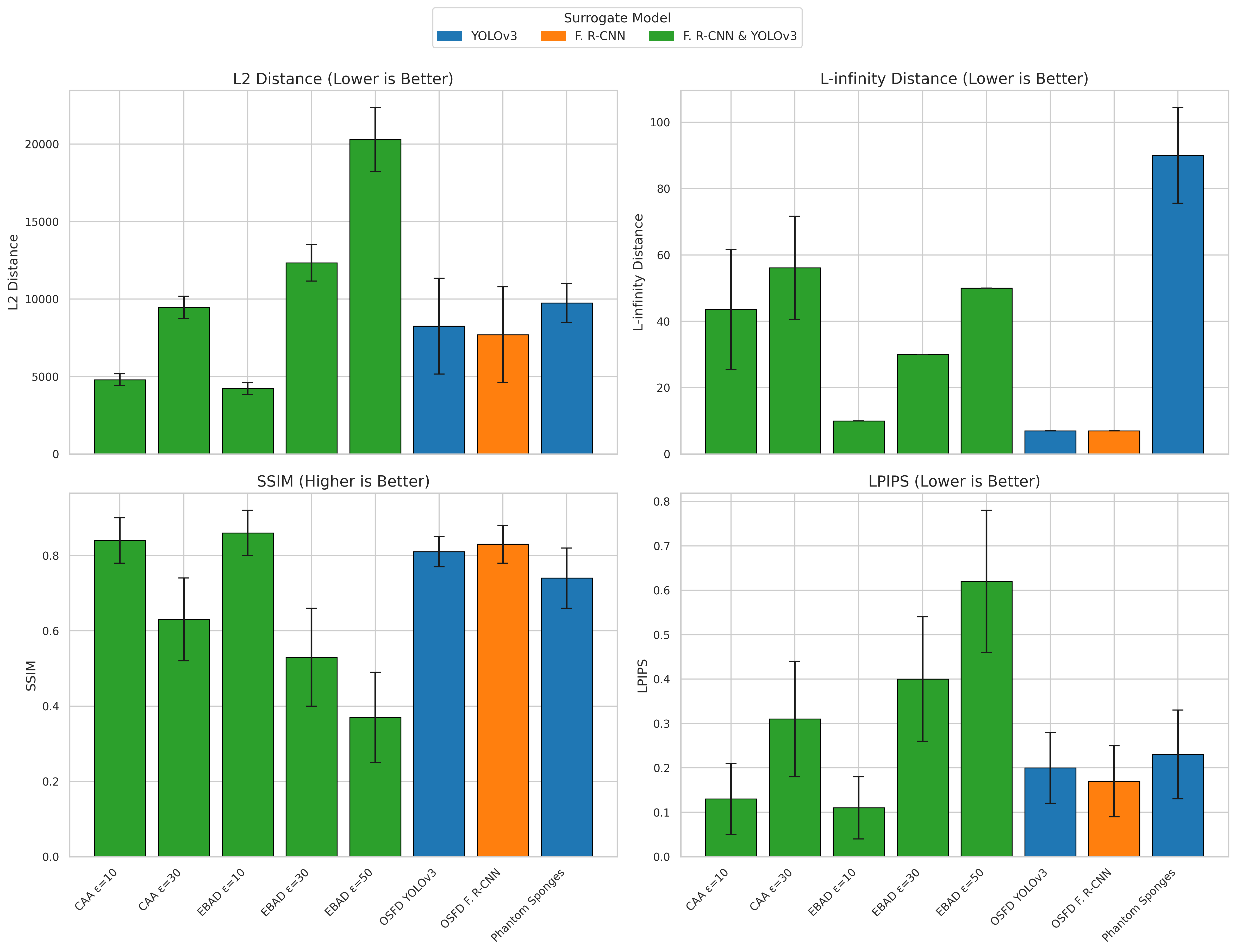}     \caption{Quantitative analysis of attack imperceptibility on the VOC2007 test set. The plots illustrate the mean and standard deviation of the perturbations across various distance ($L_2$, $L_\infty$) and perceptual (SSIM, LPIPS) metrics.}
    \label{fig:plot_dist}
\end{figure}

We now evaluate the imperceptibility of the selected attacks. For each attack, we evaluate each distance or similarity between the original and adversarial images and establish the averages and the standard deviations. Figure \ref{fig:plot_dist} plots the mean value and standard deviations for norms $L_2$ and $L_\infty$ as well as for the perceptual metrics SSIM and LPIPS. The exact values for this plot are available in Table \ref{tab:distancesbenchmark} in the appendix. These calculations are performed on non-normalized images, with RGB values between 0 and 255. For OSFD, which resizes images during its attack process, the adversarial images were resized back to their original dimensions before computing $L_2$, SSIM, and LPIPS. The $L_\infty$ norm is reported at the attack's native output resolution to reflect its $\epsilon$-constraint. For the calculation of LPIPS, we used AlexNet as in the original research paper \citep{LPIPS}.

\begin{figure}[t]
    \centering
    \includegraphics[width=\textwidth]{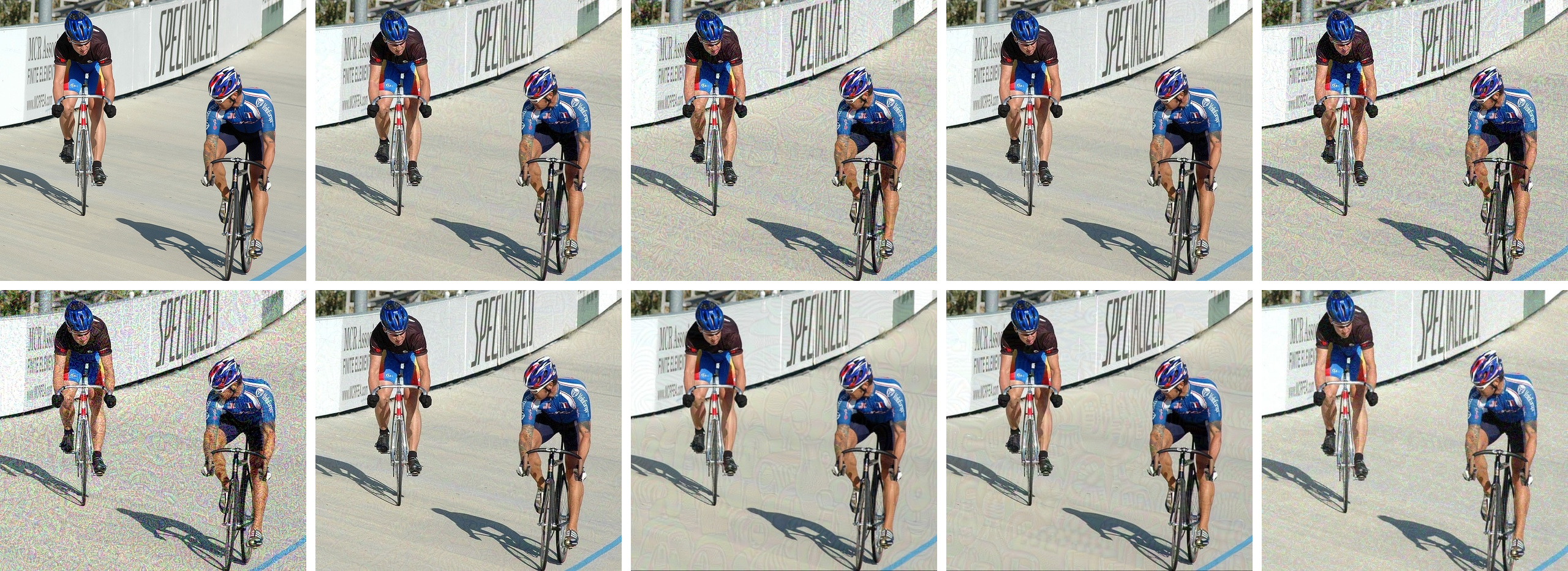}
    \caption{Visual samples from VOC2007 test set and their perturbed versions. Top row (left to right): benign image, CAA ($\epsilon=10, 30$), and EBAD on YOLOv3 ($\epsilon=10, 30$). Bottom row (left to right): EBAD on YOLOv3 ($\epsilon=50$), EBAD on Faster R-CNN ($\epsilon=10$), OSFD on YOLOv3 and Faster R-CNN ($\epsilon=5$), and PhantomSponges ($\epsilon=70$). Perturbations best viewed with zoom at 300\%.}
    \label{fig:advsamples}
\end{figure}

The problem mentioned in Section \ref{sec:criticsurvey} is clearly visualized in Figure \ref{fig:plot_dist}: The $\epsilon$ values of different methods are not comparable, as they are not applied consistently. Although $L_\infty$ is the most commonly used norm in the literature, it completely fails to model the perceptibility of an attack to the human eye. After looking at the different adversarial versions of several images, an example of which is given in Figure \ref{fig:advsamples}, the most blatant example is EBAD at $\epsilon=50$, as while its $L_\infty$ norm of 50 is high, this single number fails to capture the severe visual distortion represented by its LPIPS score of 0.62. An even more compelling case is OSFD (generated on YOLOv3): it has the lowest $L_\infty$ norm in our benchmark (7.0), yet its LPIPS score (0.19) is nearly double that of EBAD at $\epsilon=10$ (LPIPS 0.11). Based on $L_\infty$ alone, OSFD would be considered the most imperceptible attack, which is visually false (cf. Figure \ref{fig:advsamples}(h)). The original work on the LPIPS metric demonstrated that metrics based on deep features align significantly better with human perceptual judgments than traditional metrics \citep{LPIPS}. Supported by this finding we conclude that the LPIPS distance is the most appropriate metric for modeling the perceptibility of adversarial perturbations.

When comparing the LPIPS values to the traditional norms, we observe that the $L_2$ distance is more strongly correlated with LPIPS than the $L_\infty$ distance. Attacks with lower $L_2$ values consistently yield lower LPIPS scores, indicating greater imperceptibility. Conversely, attacks with higher $L_2$ values (EBAD $\epsilon=50$, CAA $\epsilon=30$) result in significantly higher LPIPS scores (0.31-0.62). This suggests that $L_2$ could be a more appropriate constraint for adversarial attacks to align the mathematical perturbation magnitude with human perceptibility.

We also notice across all attacks in Figure \ref{fig:plot_dist} the high standard deviation in LPIPS values. For example, CAA ($\epsilon=30$) shows a mean LPIPS of 0.31 with a standard deviation of 0.13, while the EBAD ($\epsilon=50$) variant shows standard deviations around 0.14-0.16. This significant variance indicates that the perceptibility of the generated perturbations is not uniform across all images within the dataset for a given attack. This highlights the limitation of evaluating perceptibility metrics across whole datasets and that perceptibility should be evaluated at the image level.

Our empirical findings suggest a strong correlation between LPIPS score and human perception. For instance, attacks with an average LPIPS distance under 0.2 are hardly perceptible to the naked eye whereas attacks with an LPIPS distance above 0.3 have visible artifacts. This observation is purely empirical and finding what level of LPIPS score corresponds to visible alterations for an adversarial attack would require an in-depth study and is beyond the scope of this paper. At an LPIPS of 0.6, images generated with the EBAD ($\epsilon=50$) variant are significantly distorted, we focus on variants of lower $\epsilon$ for EBAD in the following experiments. This is also reflected in the SSIM metric, which shows a clear inverse correlation with LPIPS (e.g., EBAD ($\epsilon=50$) has both the highest LPIPS and lowest SSIM), suggesting both are effective at identifying severe distortions.

\subsubsection{Evaluation of the attacks' impact on detector performance}

\begin{sidewaystable}
\caption{Impact of the selected attacks on detectors' localization (Relative drop in \% compared to the benign $\text{AP}_{\text{loc}}$ in \%) and classification (Relative drop in \% compared to the benign CSR in \%) abilities. The "Benign" column refers to the absence of attacks. FRC is short for Faster R-CNN. MRC is short for Mask R-CNN}
\label{tab:aploc_csr_benchmark}
\begin{center}
\begingroup
\scriptsize 
\setlength{\tabcolsep}{3.5pt} 
\renewcommand{\arraystretch}{1.3} 
\begin{tabular}{ll|c|ccccccc} 
\toprule
 & & & \multicolumn{7}{c}{\textbf{Relative drop in \% compared to the benign $\text{AP}_{\text{loc}}$ / CSR (\%)}↓} \\
\cmidrule(l){4-10}
\textbf{DETECTOR} & \textbf{Backbone} & \makecell{\textbf{Benign} \\ \textbf{$\text{AP}_{\text{loc}}$} \\ \textbf{/CSR(\%)}} & \makecell{CAA \\ (YOLO,FRC)} & \multicolumn{2}{c}{EBAD ($\epsilon=30$)} & \multicolumn{3}{c}{OSFD ($k=3$)} & \makecell{Phantom \\ Sponges}\\
\cmidrule(lr){4-4} \cmidrule(lr){5-6} \cmidrule(lr){7-9}
 & & & $\epsilon=30$ & YOLO & FRC & YOLO & FRC & MRC & \\
\midrule
        YOLOv3 & DarkNet-53 & 79.6 / 85.8 & 52.1 / \underline{67.8} & 19.5 / 51.0 & 18.1 / 40.7 & \textbf{93.5} / \textbf{70.2} & \underline{70.0} / 42.8 & 60.2 / 55.9 & 10.1 / 11.4 \\
        FCOS & ResNet-50 & 81.6 / 91.3 & 24.9 / 24.1 & 16.8 / 29.2 & 16.8 / 22.8 & \underline{95.8} / \underline{69.1} & \textbf{97.9} / \textbf{75.9} & 87.1 / 61.3 & 11.8 / 9.3 \\
        FRC & ResNet-50 & 82.7 / 89.8 & 59.7 / \textbf{80.7} & 21.3 / 52.9 & 20.1 / 43.0 & \underline{95.2} / 66.0 & \textbf{97.5} / \underline{72.9} & 91.5 / 65.9 & 11.4 / 10.0 \\
        DETR & ResNet-50 & 83.0 / 91.9 & 35.2 / 28.7 & 20.1 / 34.3 & 18.4 / 27.9 & \underline{93.9} / \underline{66.7} & \textbf{95.9} / \textbf{68.8} & 92.0 / 66.4 & 12.5 / 8.3 \\
        MRC & Swin-T & 81.6 / 91.3 & 24.9 / 24.1 & 16.8 / 29.2 & 16.8 / 22.8 & \underline{95.8} / \underline{69.1} & \textbf{97.9} / \textbf{75.9} & 87.1 / 61.3 & 11.8 / 9.3 \\
        YOLOX-l & CSPNet & 89.5 / 94.2 & 7.7 / 7.5 & 8.4 / 10.7 & 7.4 / 8.8 & \textbf{89.3} / \textbf{63.4} & 74.3 / 43.0 & \underline{76.3} / \underline{55.4} & 13.7 / 11.1 \\
        DINO & Swin-L & 89.9 / 96.9 & 2.8 / 2.3 & 3.6 / 2.8 & 3.6 / 2.3 & \underline{29.9} / \underline{12.4} & 16.9 / 5.2 & \textbf{43.4} / \textbf{18.5} & 1.7 / 0.9 \\
\bottomrule
\end{tabular}
\endgroup
\end{center}
\end{sidewaystable}

\begin{table}[t]
\caption{Impact of the selected attacks on detectors' performance (Relative drop in \% compared to the benign mAP in \%). The "Benign" column refers to the absence of attacks. The final row shows the average attack generation time per image on a NVIDIA TITAN RTX. FRC is short for Faster R-CNN. MRC is short for Mask R-CNN}
\label{tab:mapbenchmark}
\centering 
\begingroup
\footnotesize 
\setlength{\tabcolsep}{3.5pt} 
\renewcommand{\arraystretch}{1.3} 
\begin{tabular}{ll|c|cccccccc} 
\toprule
 & & & \multicolumn{8}{c}{\textbf{Relative drop in \% compared to the benign mAP ↓}} \\
\cmidrule(l){4-11}
\textbf{DETECTOR} & \textbf{Backbone} & \makecell{\textbf{Benign} \\ \textbf{mAP}} & \multicolumn{2}{c}{\makecell{CAA \\ (YOLO,FRC)}} & \multicolumn{2}{c}{EBAD ($\epsilon=30$)} & \multicolumn{3}{c}{OSFD ($k=3$)} & \makecell{Phantom \\ Sponges}\\
\cmidrule(lr){4-5} \cmidrule(lr){6-7} \cmidrule(lr){8-10}
 & & & $\epsilon=10$ & $\epsilon=30$ & YOLO & FRC & YOLO & FRC & MRC & \\
\midrule
        YOLOv3 & DarkNet-53 & 75.8 & 83.8 & \textbf{92.1} & 74.5 & 69.3 & \underline{91.3} & 70.4 & 67.3 & 16.0 \\ 
        FCOS & ResNet-50 & 78.9 & 17.7 & 41.2 & 39.0 & 36.0 & \underline{92.6} & \textbf{96.7} & 88.7 & 18.8 \\ 
        FRC & ResNet-50 & 79.3 & 89.0 & 95.7 & 72.5 & 67.1 & \underline{90.7} & \textbf{95.7} & 89.8 & 17.2 \\ 
        DETR & ResNet-50 & 80.3 & 23.9 & 54.7 & 53.9 & 47.4 & 89.5 & \textbf{93.4} & \underline{89.8} & 18.4 \\
        MRC & Swin-T & 86.9 & 5.6 & 15.8 & 17.1 & 16.7 & \underline{85.5} & 77.6 & \textbf{99.4} & 10.9 \\ 
        YOLOX-l & CSPNet & 87.5 & 3.8 & 10.6 & 12.1 & 10.9 & \underline{84.3} & 70.2 & \textbf{76.1} & 19.1 \\
        DINO & Swin-L & 89.6 & 0.7 & 3.8 & 4.5 & 4.4 & \underline{27.3} & 15.4 & \textbf{40.2} & 2.3 \\
\midrule
\multicolumn{2}{l|}{\textbf{Time per image (s)}} & - & \multicolumn{2}{c}{84 s} & \multicolumn{2}{c}{17 s} & \multicolumn{3}{c}{44 s} & \makecell{N/A \\ (Universal)} \\
\bottomrule 
\end{tabular}
\endgroup
\end{table}

Tables \ref{tab:mapbenchmark} and \ref{tab:aploc_csr_benchmark} show, respectively, the mAP, $\text{AP}_\text{loc}$ and CSR values of the detectors under adversarial attack. We can derive the impact of each attack, with regard to their different outcomes and their black-box transferability across all detectors.

\paragraph{Type of impact:} 
As we expect, the two mislabeling attacks, CAA and EBAD, particularly affect CSR. Because CAA can sometimes cause the appearance of spurious detections by modifying the context of the images, its impact on localization is greater than EBAD \citep{CAA}. On the other hand, EBAD focuses solely on mislabeling and therefore has less impact on localization than most other attacks \citep{EBAD}; for instance, on YOLOv3, EBAD (YOLO) causes a 51.0\% drop in CSR compared to only 19.5\% in $\text{AP}_{\text{loc}}$ (cf. Table \ref{tab:aploc_csr_benchmark}). The OSFD attack, being the newest, is the most effective of the selected attacks in terms of mAP. As an attack with random output, it targets both localization and classification. This is reflected in the $\text{AP}_\text{loc}$ and CSR results which are impressive, especially for $\text{AP}_\text{loc}$ (drop of more than 90\% for every model except DINO for OSFD (YOLO)). PhantomSponges is a universal attack and, as expected, is therefore less effective than the others. As the objective of PhantomSponges is to generate new non-existent objects while preserving the original detections, its effect is deliberately very weak on CSR and is less severe than the others on $\text{AP}_\text{loc}$ ($\sim$10-11\% drop) \citep{PhantomSponges}.

\paragraph{Transferability:} 
The most effective attack on a large number of detectors is OSFD, which is also the most recent and based on feature manipulation. This is a random output attack that achieves high transferability while remaining highly effective on all detectors tested in white-box and black-box configurations causing mAP drops exceeding 84\% on all CNN-based models (Table \ref{tab:mapbenchmark}). It is worth noting that this efficiency comes at a significant computational cost. As shown in the final row of Table \ref{tab:mapbenchmark}, OSFD requires an average of 44 seconds per image in our benchmark, making it considerably slower than the highly-efficient EBAD (17 s) but faster than the context-aware CAA (84 s). For the other attacks, we remark that CAA is more efficient in a white-box setting (e.g., 95.7\% drop on FRC vs 67.1\% drop on FRC for EBAD) compared to EBAD that exhibits more transferability in a black-box setting (e.g., 12.1\% drop on YOLOX vs 10.6\% for CAA).

In terms of \textbf{detectors}, the most recent models such as YOLOX \citep{YOLOX} or DINO \citep{dino} exhibit better robustness. This can partially be attributed to the fact that they are always tested in black-box configuration and without ever being part of the surrogate models EBAD, CAA and OSFD. For EBAD and OSFD, it is the attacks generated on YOLOv3 that achieve better transferability across detectors except for DINO. Another interesting finding is the existence of a significant gap in transferability between \textbf{architectures}, particularly from Convolutional Neural Networks (CNNs) to Vision Transformers. We can also note the robustness of DINO, which consistently maintains high performance  against attacks transferred from other models, restricting the mAP drop to 27.3\% for OSFD and under 5\% for other attacks (Table \ref{tab:mapbenchmark}). As there are currently no attacks in our benchmark capable of compromising modern detectors in a realistic black-box scenario, future research must focus on formulating new black-box attacks designed to challenge these modern architectures.

\section{How to defend object detectors against the most efficient attacks?}
\label{sec:defend}
\subsection{Performing adversarial training}

Although the field of adversarial attacks has progressed a lot in recent years, the same cannot be said for defense methods, which are much less numerous, as shown in Section \ref{sec:reviewDEF}. To our knowledge, adversarial training is currently the defense method with the greatest potential to make detectors robust against the largest number of attacks at once \citep{AdvtrainGabor}. A key objective of this work is to systematically evaluate which attacks are most suitable for generating effective training data. In order to make the best use of this method, we fine-tune our detectors using images perturbed by state-of-the-art attacks. We are also mixing adversarial images from different attacks in the adversarial training dataset, hoping to observe a positive effect and thus making the detector even more robust. The goal is to identify effective adversarial training strategies (which attacks/combinations) to maximize detector robustness while maintaining performance on benign images. 

For this experiment, we select the same attacks as in Section \ref{sec:benchmark}. The adversarial examples are generated in a white-box configuration on the tested detector (Faster R-CNN or YOLOv3) as this ensures the attack is specifically tailored to the model we are defending and represents its "worst-case" scenario. There are two exceptions: CAA does not consider a victim model \citep{CAA}, and PhantomSponges, which is always generated with a YOLOv3 \citep{PhantomSponges}, as explained in Section \ref{sec:choice}.

We include attack configurations that may lead to more perceptible perturbations (as discussed in Section \ref{sec:benchmark}) because such stronger adversarial examples are valuable for potentially inducing greater robustness during the adversarial training process.

Comprehensive data is provided in the Appendix, including the full comparison for both YOLOv3 and Faster R-CNN (Tables \ref{tab:mapyoloadvtrainannexe} and \ref{tab:mapfrcnnadvtrainannexe}) as well as the detailed metric breakdowns ($\text{mAP}$, $\text{AP}_{\text{loc}}$, and $\text{CSR}$) for the Faster R-CNN experiments (Tables \ref{tab:mapfrcnnadvtrain}, \ref{tab:aplocfrcnnadvtrain}, and \ref{tab:csrfrcnnadvtrain}).

\subsection{Experimental setup}

In this section, we use Faster R-CNN \citep{faster-r-cnn} and YOLOv3 \citep{yolov3} introduced in Section \ref{sec:3expesetup}. To perform the adversarial training, we still use the MMDetection framework \citep{mmdetection}. Each adversarial training process is performed in the same way. As explained, the two detectors were pre-trained on COCO \citep{COCO} and are now being fine-tuned on the adversarially attacked VOC2007 trainval set \citep{VOC2007}. The learning rate is always set to $0.0001$ and the number of epochs is set to 50 for YOLOv3 and 20 for Faster R-CNN. As in Section \ref{sec:benchmark}, the test dataset is VOC2007 test set \citep{VOC2007}. We still use the IoU threshold of 0.5 for mAP, $\text{AP}_{\text{loc}}$ and CSR. When using a mixed adversarial training dataset, the images altered by the different attacks come from original images that are always different; there are never two adversarial versions of the same image in a single adversarial training set ensuring that the training set benefits from a wider diversity of images.

\subsection{Evaluation of the robustness of detectors after adversarial training}

We fine-tune each detector using adversarial images of several attacks and then we test them against the same set of attacks as in Section \ref{sec:benchmark}. For comparison, we also test the "Baseline" detector, i.e., without adversarial training, and the "Benign" detector, i.e. fine-tuned on unattacked images. For each attack, the same metrics  show its impacts on defended models, as discussed in Section \ref{sec:benchmark}.

\subsubsection{Comparison of the impact of each attack on defense by adversarial training}

\begin{sidewaystable}
\caption{Performance (\textbf{mAP} in \%) of \textbf{YOLOv3} detectors defended by adversarial training. The subscript next to CAA and EBAD indicates their respective $\epsilon$ parameters. For EBAD attacks, the target model is YOLOv3. For OSFD and Phantom Sponges, the model mentioned is the surrogate used for attack generation. Higher is better.}
\label{tab:mapyoloadvtrain}
\centering
\renewcommand{\arraystretch}{1.4} 
\begin{tabular}{c|ccccccccc} 
\toprule
\makecell{FINE-TUNED\\ON} & Benign & \makecell{$\text{CAA}_{10}$} & \makecell{$\text{CAA}_{30}$} & \makecell{$\text{EBAD}_{10}$} & \makecell{$\text{EBAD}_{30}$} & \makecell{$\text{EBAD}_{50}$} & \makecell{OSFD} & \makecell{OSFD} & \makecell{Phantom\\ Sponges}\\
& & & & & & & YOLOv3 & F. R-CNN & YOLOv3 \\
\cmidrule{2-10}
Baseline & \textbf{75.8} & 12.3 & 6.0 & 22.0 & 19.3 & 18.1 & 6.6 & 22.4 & 63.7 \\
Benign & \textbf{75.8} & 28.0 & 17.1 & 48.7 & 32.7 & 28.1 & 8.3 & 23.3 & 63.9 \\
$\text{CAA}_{30}$ & 72.2 & \underline{72.2} & \underline{66.3} & \underline{72.1} & \underline{67.4} & \underline{60.6} & \underline{53.0} & \underline{61.6} & \underline{66.8} \\
$\text{EBAD}_{10}$ & \underline{75.4} & 67.0 & 38.1 & 70.1 & 51.8 & 39.3 & 36.9 & 53.2 & 66.2 \\
$\text{EBAD}_{50}$ & 68.7 & 69.4 & \textbf{66.8} & 70.0 & \textbf{68.8} & \textbf{66.0} & 49.4 & 57.7 & 64.7 \\
OSFD & 73.5 & \textbf{73.9} & 65.1 & \textbf{73.6} & 65.2 & 53.5 & \textbf{69.1} & \textbf{69.8} & 65.4 \\
\makecell{Phantom Sponges} & 74.1 & 62.6 & 35.4 & 67.6 & 48.0 & 36.2 & 27.3 & 45.2 & \textbf{73.0} \\
\botrule
\end{tabular}
\end{sidewaystable}

\begin{sidewaystable}
\caption{Localization (AP\textsubscript{loc} in \%) and classification performance (CSR in \%) of \textbf{YOLOv3} detectors defended by adversarial training. The subscript next to CAA and EBAD indicates their respective $\epsilon$ parameters. For EBAD attacks, the target model is YOLOv3. For OSFD and Phantom Sponges, the model mentioned is the surrogate used for attack generation.  Higher is better.}
\label{tab:aploc_csr_yoloadvtrain}
\centering
\renewcommand{\arraystretch}{1.4} 
\begin{tabular}{c|ccccccccc}
\toprule
\makecell{FINE-TUNED\\ON} & Benign & \makecell{$\text{CAA}_{10}$} & \makecell{$\text{CAA}_{30}$} & \makecell{$\text{EBAD}_{10}$} & \makecell{$\text{EBAD}_{30}$} & \makecell{$\text{EBAD}_{50}$} & \makecell{OSFD} & \makecell{OSFD} & \makecell{Phantom\\ Sponges}\\
& & & & & & & YOLOv3 & F. R-CNN & YOLOv3 \\
\cmidrule{2-10}
Baseline & \underline{79.6} / \textbf{85.8} & 41.6 / 37.6 & 38.1 / 27.6 & 64.7 / 42.1 & 64.1 / 42.0 & 64.0 / 41.7 & 5.2 / 25.6 & 23.9 / 49.1 & 71.6 / 76.0 \\
Benign & \textbf{80.0} / 83.8 & 49.7 / 50.4 & 42.9 / 42.9 & 65.6 / 60.9 & 63.5 / 49.7 & 62.9 / 46.2 & 5.2 / 22.3 & 24.1 / 43.3 & 70.2 / 72.3 \\
$\text{CAA}_{30}$ & 77.5 / 82.1 & \underline{77.1} / \underline{80.5} & \textbf{73.7} / \underline{75.0} & \underline{76.5} / \underline{79.4} & \underline{72.9} / \underline{74.5} & \underline{69.4} / \underline{68.4} & \underline{60.0} / \underline{68.2} & \underline{68.2} / \underline{74.4} & \underline{72.7} / \underline{77.4} \\
$\text{EBAD}_{10}$ & \underline{79.6} / \underline{84.4} & 73.8 / 76.2 & 57.2 / 59.4 & 74.8 / 76.9 & 67.6 / 63.1 & 65.2 / 54.6 & 39.8 / 59.6 & 58.7 / 71.0 & 72.6 / 76.5 \\
$\text{EBAD}_{50}$ & 74.5 / 79.9 & 74.9 / 79.4 & \underline{73.4} / \textbf{76.7} & 75.1 / 78.5 & \textbf{74.1} / \textbf{76.5} & \textbf{72.8} / \textbf{73.9} & 57.5 / 66.4 & 65.2 / 72.6 & 71.1 / 77.1 \\
OSFD & 77.7 / 82.2 & \textbf{77.7} / \textbf{81.3} & 72.3 / 73.0 & \textbf{77.3} / \textbf{80.5} & 71.6 / 72.0 & 66.6 / 62.7 & \textbf{73.8} / \textbf{78.1} & \textbf{74.6} / \textbf{79.3} & 70.9 / 74.5 \\
\makecell{Phantom Sponges} & 78.7 / 83.3 & 71.1 / 73.1 & 55.1 / 56.7 & 73.7 / 75.0 & 66.7 / 60.7 & 64.5 / 52.3 & 33.7 / 49.0 & 52.2 / 64.2 & \textbf{77.0} / \textbf{81.2} \\
\botrule
\end{tabular}
\end{sidewaystable}

We observe (cf. Tables \ref{tab:mapyoloadvtrain} and \ref{tab:aploc_csr_yoloadvtrain}) that even fine-tuning on benign images almost always improves the robustness of the detectors, particularly against mislabeling attacks (e.g., mAP against EBAD $\epsilon =10$ improves from 22.0\% to 48.7\%). Given that our detectors are trained on COCO and then fine-tuned and tested on subsets of VOC2007, we can conclude that the reduction of the domain gap for the test makes the detector more robust against this type of perturbation. Indeed, this effect is more noticeable with mAP and CSR than with $\text{AP}_{\text{loc}}$.

Table \ref{tab:mapyoloadvtrain} also clearly illustrates the expected robustness-accuracy trade-off: For adversarial training made on EBAD, the more the magnitude of the perturbations increases, the more the performance on adversarial images increases (e.g., against CAA $\epsilon=30$, mAP rises from 38.1\% to 66.8\%) at the expense of performance on original images (benign mAP drops from 75.4\% to 68.7\%). This confirms that achieving broad robustness requires accepting a compromise in benign performance.

Do specific attacks provide superior robustness? OSFD and CAA lead to the most robust detectors overall. This points to a critical result: the robustness gained from training on one attack does not necessarily generalize to all others. Specifically, robustness does not transfer well from weaker to stronger attacks. This is most evident when testing against OSFD: Models trained on CAA $\epsilon=30$ or EBAD $\epsilon=50$ are still vulnerable to OSFD (mAP of 53.0\% and 49.4\%, respectively). However, the reverse is not true: the OSFD-trained model performs well against CAA and EBAD (mAP >65\% against both).

Similarly, using a universal perturbation such as PhantomSponges for adversarial training leads to weaker improvement as expected from a unique perturbation.

\subsubsection{Adversarial training on partially attacked datasets}

\begin{figure}[t]
    \centering
    \includegraphics[width=1\textwidth]{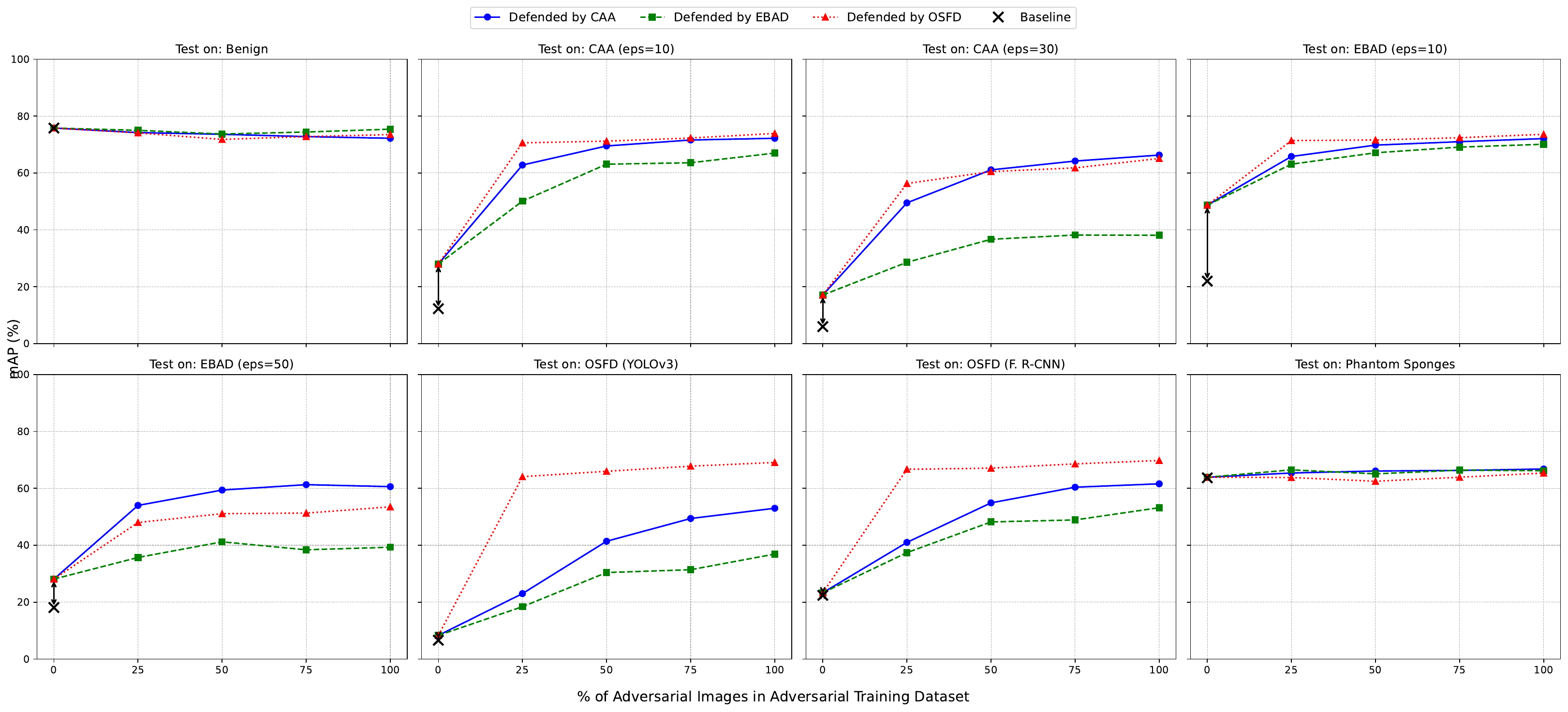}
    \caption{Performance (mAP in \%) of YOLOv3 fine-tuned with varying percentages of adversarial images produced by three different attacks ($\text{CAA}_{30}$, $\text{EBAD}_{10}$ and OSFD (YOLO)) and tested across eight different adversarial attacks.}
    \label{fig:partial_plots_yolo}
\end{figure}

To find an optimal accuracy-robustness trade-off, this experiment investigates if an equilibrium can be found by performing adversarial training on datasets containing only a proportion of adversarial images mixed with benign ones. We fine-tune YOLOv3 on training sets composed of 25\%, 50\%, and 75\% adversarial images (generated by CAA $\epsilon=30$, EBAD $\epsilon=10$ and OSFD (YOLO)), with the remainder of the dataset being benign images. We then evaluate these fine-tuned models against models trained on 100\% adversarial images.

The observed results in Figure \ref{fig:partial_plots_yolo} show that the performance gain on unattacked images is often low compared to the performance drop on adversarial images. For example, the OSFD-trained model's performance only degrades from 75.8\% mAP (at 0\% adversarial) to 73.5\% mAP (at 100\% adversarial). This minor 2.3 percentage point (p.p.) loss in benign accuracy is dramatically outweighed by the massive gain in robustness across all attacks. This trade-off is particularly poor at 25\% of attacked images in the adversarial training dataset. For example, the robustness of the CAA-trained model against OSFD (YOLO) plummets from 53.0\% mAP (at 100\% adversarial) to just 23.0\% mAP (at 25\% adversarial) for a recovery of 2\% mAP on the benign dataset. These results suggest that conducting adversarial training with a fully attacked dataset is likely the more efficient strategy.

Our results show that while OSFD provides the best general robustness, it still leaves the model vulnerable to high-magnitude attacks like EBAD $\epsilon=50$.  Similarly, EBAD $\epsilon=50$ training confers strong robustness against mislabeling but fails to protect against OSFD. This suggests that mixing attacks with different outcomes (random output for OSFD and mislabeling for EBAD) could cover these complementary weaknesses.

\subsubsection{Adversarial training on mixed-attack datasets}

\begin{sidewaystable}
\caption{Performance (mAP in \%) of \textbf{YOLOv3} detectors defended by adversarial training with various mixed dataset compositions. The subscript next to CAA and EBAD indicates their respective $\epsilon$ parameters. For EBAD attacks, the target model is YOLOv3. For OSFD and Phantom Sponges, the model mentioned is the surrogate used for attack generation. Bold and underlined are respectively best and second best YOLO defended models. Higher is better.}
\label{tab:yolomapadvtrain_mixed}
\centering
\renewcommand{\arraystretch}{1.4} 
\begin{tabular}{c|ccccccccc} 
\toprule
\makecell{FINE-TUNED\\ON} & Benign & \makecell{CAA\\$\epsilon=10$} & \makecell{CAA\\$\epsilon=30$} & \makecell{EBAD\\$\epsilon=10$} & \makecell{EBAD\\$\epsilon=30$} & \makecell{EBAD\\$\epsilon=50$} & \makecell{OSFD} & \makecell{OSFD} & \makecell{Phantom\\ Sponges}\\
& & & & YOLOv3 & YOLOv3 & YOLOv3 & YOLOv3 & F. R-CNN & YOLOv3 \\
\cmidrule{2-10}
Benign & \textbf{75.8} & 28.0 & 17.1 & 48.7 & 32.7 & 28.1 & 8.3 & 23.3 & 63.9 \\
$\text{CAA}_{30}$ & 72.2 & 72.2 & 66.3 & 72.1 & 67.4 & 60.6 & 53.0 & 61.6 & 66.8 \\
OSFD & \underline{73.5} & \textbf{73.9} & 65.1 & \textbf{73.6} & 65.2 & 53.5 & \textbf{69.1} & \textbf{69.8} & 65.4 \\
$\text{EBAD}_{50}$ & 68.7 & 69.4 & 66.8 & 70.0 & 68.8 & \underline{66.0} & 49.4 & 57.7 & 64.7 \\
\makecell{OSFD 50\% $\text{CAA}_{30}$ 50\%} & 72.2 & 72.9 & 67.6 & 72.8 & 67.6 & 59.5 & 67.2 & 68.5 & 66.8 \\
\makecell{OSFD 75\% $\text{CAA}_{30}$ 25\%} & 73.1 & 73.5 & 67.5 & \textbf{73.6} & 67.6 & 58.4 & 68.2 & \underline{69.3} & \textbf{67.4} \\
\makecell{OSFD 33\% $\text{CAA}_{30}$ 33\% \\ Benign 34\%} & 71.7 & 72.1 & 65.6 & 71.8 & 67.0 & 59.0 & 65.9 & 67.4 & 66.0 \\
\makecell{OSFD 50\% $\text{EBAD}_{50}$ 50\%} & 71.9 & 72.6 & \underline{69.5} & 73.0 & 69.7 & 65.2 & 67.2 & 68.7 & \underline{67.1} \\
\makecell{OSFD 25\% $\text{EBAD}_{50}$ 75\%} & 70.8 & 71.8 & \underline{69.5} & 71.9 & \underline{69.9} & \textbf{66.2} & 65.5 & 67.0 & 65.7 \\
\makecell{OSFD 75\% $\text{EBAD}_{50}$ 25\%} & 73.0 & \underline{73.7} & \textbf{69.6} & \underline{73.5} & \textbf{70.0} & 64.8 & \underline{68.9} & \textbf{69.8} & \textbf{67.4} \\
\botrule
\end{tabular}
\end{sidewaystable}

\begin{sidewaystable}
\caption{Localization (AP\textsubscript{loc} in \%) and classification performance (CSR in \%) of \textbf{YOLOv3} detectors defended by adversarial training with various mixed dataset compositions. The subscript next to CAA and EBAD indicates their respective $\epsilon$ parameters. For EBAD attacks, the target model is YOLOv3. For OSFD and Phantom Sponges, the model mentioned is the surrogate used for attack generation. Bold and underlined are respectively best and second best YOLO defended models. Higher is better.}
\label{tab:yolomap_csr_advtrain_mixed}
\centering
\renewcommand{\arraystretch}{1.7} 
\begin{tabular}{c|ccccccccc} 
\toprule
\makecell{FINE-TUNED\\ON} & Benign & \makecell{CAA\\$\epsilon=10$} & \makecell{CAA\\$\epsilon=30$} & \makecell{EBAD\\$\epsilon=10$} & \makecell{EBAD\\$\epsilon=30$} & \makecell{EBAD\\$\epsilon=50$} & \makecell{OSFD} & \makecell{OSFD} & \makecell{Phantom\\ Sponges}\\
& & & & YOLOv3 & YOLOv3 & YOLOv3 & YOLOv3 & F. R-CNN & YOLOv3 \\
\cmidrule{2-10}
Benign & \textbf{80.0}/\textbf{83.8} & 49.7/50.4 & 42.9/42.9 & 65.6/60.9 & 63.5/49.7 & 62.9/46.2 & 5.2/22.3 & 24.1/43.3 & 70.2/72.3 \\
$\text{CAA}_{30}$ & 77.5/82.1 & 77.1/80.5 & 73.7/75.0 & 76.5/79.4 & 72.9/74.5 & 69.4/68.4 & 60.0/68.2 & 68.2/74.4 & \textbf{72.7}/77.4 \\
OSFD & \underline{77.7}/\underline{82.2} & \textbf{77.7}/\underline{81.3} & 72.3/73.0 & \underline{77.3}/\underline{80.5} & 71.6/72.0 & 66.6/62.7 & \textbf{73.8}/\textbf{78.1} & \textbf{74.6}/\textbf{79.3} & 70.9/74.5 \\
$\text{EBAD}_{50}$ & 74.5/79.9 & 74.9/79.4 & 73.4/76.7 & 75.1/78.5 & 74.1/76.5 & \underline{72.8}/\underline{73.9} & 57.5/66.4 & 65.2/72.6 & 71.1/77.1 \\
\makecell{OSFD 50\% $\text{CAA}_{30}$ 50\%} & 76.8/81.6 & 77.4/\textbf{81.4} & 73.9/76.1 & 77.0/80.3 & 73.2/74.8 & 69.4/68.3 & 72.4/76.8 & 73.3/77.9 & \textbf{72.7}/76.9 \\
\makecell{OSFD 75\% $\text{CAA}_{30}$ 25\%} & 77.3/81.2 & \underline{77.6}/81.2 & 74.2/75.4 & \textbf{77.4}/80.3 & 73.0/74.0 & 69.1/66.8 & 73.0/77.1 & 73.9/78.2 & 72.4/76.7 \\
\makecell{OSFD 33\% $\text{CAA}_{30}$ 33\% \\ Benign 34\%} & 76.3/79.4 & 76.5/79.1 & 72.9/73.5 & 76.2/78.2 & 72.5/73.4 & 69.4/67.8 & 71.3/74.8 & 72.6/75.8 & 71.8/74.9 \\
\makecell{OSFD 50\% $\text{EBAD}_{50}$ 50\%} & 76.7/81.5 & 77.2/\underline{81.3} & \textbf{75.5}/\underline{77.8} & 77.2/\textbf{80.6} & \underline{74.7}/\underline{77.1} & 72.5/73.7 & 72.6/77.3 & 73.6/78.5 & \underline{72.6}/\textbf{77.8} \\
\makecell{OSFD 25\% $\text{EBAD}_{50}$ 75\%} & 77.2/81.9 & \textbf{77.7}/\underline{81.3} & \underline{75.1}/77.5 & \underline{77.3}/\textbf{80.6} & \underline{74.7}/76.7 & 72.0/72.6 & \underline{73.3}/\underline{77.8} & \underline{74.2}/\underline{78.7} & \textbf{72.7}/\underline{77.5} \\
\makecell{OSFD 75\% $\text{EBAD}_{50}$ 25\%} & 75.8/80.5 & 76.7/80.7 & \underline{75.4}/\textbf{78.3} & 76.6/79.7 & \textbf{75.1}/\textbf{77.2} & \textbf{73.1}/\textbf{74.2} & 71.4/75.9 & 72.5/77.1 & 72.1/76.7 \\
\botrule
\end{tabular}
\end{sidewaystable}

We first test mixing the two best-performing single attacks, OSFD and CAA $\epsilon=30$ (Tables \ref{tab:yolomapadvtrain_mixed} and \ref{tab:yolomap_csr_advtrain_mixed}) in different proportions. The results show a minor benefit: the $\text{OSFD 75\% CAA}_{30} \text{ 25\%}$ mix achieves a slightly better mAP against EBAD $\epsilon=50$ (58.4\%) than OSFD alone (53.5\%). Next, we test mixing OSFD with EBAD $\epsilon=50$. The results show a clear synergistic effect. The $\text{OSFD 75\% EBAD}_{50} \text{ 25\%}$ model, for instance, produces a detector that is superior to the OSFD-only model. While it retains the robustness of OSFD (e.g., 68.9\% mAP vs. OSFD (YOLO)), it achieves 64.8\% mAP against EBAD $\epsilon=50$, a massive improvement over the 53.5\% mAP from the OSFD-only model. We also tested adding 34\% of unattacked images to this mix, but it does not provide many clear benefits.

These mixtures demonstrate the positive effect that can be obtained for adversarial training by using attacks with different outcomes, especially if these outcomes are very different, such as with random output (OSFD) and object mislabeling (EBAD) used here. This indicates that the training forces the model to learn more robust features against spatial (localization) and semantic (classification) perturbations.

\section{Future directions}

\subsection{Bridging the Cross-Architecture Transferability Attack Gap}

There is a notable gap in the literature regarding adversarial attacks generated on newer or different detectors. This includes more up-to-date versions of YOLO or end-to-end transformer architectures such as DINO. So far, few attacks have used the latter to generate their perturbations \citep{EBAD} (cf. Section \ref{sec:benchmark}). This is largely because most existing attack codebases rely on older frameworks (e.g., earlier versions of MMDetection for EBAD, CAA and OSFD) that do not include these modern detectors. Furthermore, even recent attacks that do include these modern detectors, such as AFOG \citep{AFOG}, often do not focus on cross-architecture transferability, which highlights a blind spot in current attack methodologies.

Beyond targeting the output of detectors, attacks that focus on high-level representations to generate perturbations like OSFD \citep{OSFD} yield promising results. Feature-based attacks could be explored as a way of formulating attacks with better cross-architecture transferability \citep{OSFD} (cf. Section \ref{sec:benchmark}). \\
A special focus should be made around the transferability of black-box attacks for modern detectors. Our findings show that attacks seem to be less transferable for modern detectors, especially cross-architecture. Just as we have done in our benchmark, we hope that future attack methodologies will be consistently evaluated in a black-box scenario against a panel of modern detectors to better assess their transferability.

\subsection{Adversarial training and new defense strategies}
There are still various aspects of adversarial training that need to be explored. While this work focused on cross-detector transferability, a critical open question remains regarding cross-dataset transferability. In real-world deployment, the attack environment often differs significantly from the training domain (e.g., different weather conditions, lighting, or sensor types) \citep{DomainVar}. It remains unclear how well adversarial training on a standard benchmark generalizes to these out-of-distribution scenarios. A significant open question in the literature is whether the robustness gained from adversarial training is domain-specific or if it can help bridge the domain gap between disparate environments \citep{TranferRobust}.

Research is also needed for new defense methods (cf. Section \ref{sec:reviewDEF}). At the moment, the number of methods is rather limited, but progress could be made. Drawing inspiration from classification, important improvements could be made with methods such as Adversarial purification \citep{AdvDiff} , which attempts to remove perturbations before the image is fed to the model with the use of generative models. Another possibility would be to improve adversarial training by including other methods such as Tradeoff-inspired Adversarial Defense via Surrogate-loss minimization (TRADES), which focuses on the fact that improving robustness against adversarial attacks often comes at the cost of accuracy. They focus on optimizing the original error while minimizing the distance between predictions of original and adversarial examples \citep{TRADES}.

\subsection{Standard framework for patch-based attacks}

Our observations about the lack of standard evaluation for attacks for object detection can be extended to patch-based attacks  \citep{ImpBackPatches, RPAttack} and physical attacks \citep{InvCloak, AdvTshirt} even though the literature about patch-based attacks for object detection is quite developed. A key question is how can we design a standard benchmark and metrics for this distinct threat model? For example, for patch-based attacks, the perceptibility of a patch is related to factors like patch size or the semantic consistency with its environment which are measured differently than standard $L_p$ distances. Similarly, physical attacks and adversarial objects must remain effective even when subjected to a wide range of natural transformations that do not exist in the purely digital domain, such as changing lighting conditions, different viewing angles, and partial occlusions, etc. 

\section{Conclusion}

In this paper, we addressed the critical lack of standardized evaluation in adversarial attack and defense of object detectors. We provided a review of adversarial attacks and defenses in object detection that identified the disparate datasets, inconsistent metrics, and non-comparable perturbation measures, making fair comparisons impossible. To resolve this fragmentation, we proposed a unified benchmark framework focusing on digital, non-patch-based attacks. By enforcing consistent datasets, metrics, and perturbation constraints, our framework enables the first direct comparison of these methods. This benchmark introduces specialized metrics ($\text{AP}_{\text{loc}}$ and CSR) to disentangle localization and classification errors and leverages perceptual metrics (LPIPS, $L_2$) to provide a fair measure of attack cost or perceptibility.

Using this benchmark, our comparison of SOTA attack performance led to several key findings. We identified OSFD as the most effective and broadly transferable attack, but highlighted its significant computational cost (44s per image). Our analysis also reveals the crucial compromises with perceptibility, demonstrating that the $L_\infty$ norm is a poor proxy for human perception and that perceptual metrics like LPIPS are essential for fair comparisons. Critically, we found that black-box transferability is not uniform: attacks generated on YOLOv3 (like EBAD and OSFD) transfer well to other CNNs, but we revealed a significant cross-architectural robustness gap, as all tested attacks fail to transfer to modern transformer-based architectures like DINO. On the defense side, our experiments on adversarial training identified clear strategies. We demonstrated that using a 100\% adversarially-attacked dataset is superior to mixing with benign images, as the minimal loss in benign accuracy is a worthwhile trade-off for the substantial gain in robustness. Most importantly, we found that the strongest defense is achieved by training on a mix of high-perturbation, complementary attacks (e.g., spatial and semantic), which successfully covers the weaknesses of any single, strong attack and achieves the highest overall robustness. 

Our work provides this clear comparison of modern attacks and defenses, establishing best practices for adversarial defense and identifying the cross-architecture robustness of transformers as the next major frontier for both attackers and defenders.

\newpage 

\backmatter

\section*{Declarations}

\subsection*{Author information}
\begin{itemize}
    \item Alexis Winter: \url{https://orcid.org/0009-0000-3006-525X}
    \item Romaric Audigier: \url{https://orcid.org/0000-0002-4757-2052}
    \item Angelique Loesch: \url{https://orcid.org/0000-0001-5427-3010}
    \item Bertrand Luvison: \url{https://orcid.org/0000-0003-2475-3712}
\end{itemize}

\subsection*{Funding}
Funded by the European Union. Views and opinions expressed are however those of the author(s) only and do not necessarily reflect those of the European Union nor the European Commission. Neither the European Union nor the granting authority can be held responsible for them. This work was supported under the EDF Project FaRADAI (grant number 101103386).
This work was made possible by the FactoryIA supercomputer (funded by the Ile-de-France Regional Council).
This work was supported by a State grant managed by the French National Research Agency, Agence Nationale de la Recherche, under "France 2030" (grant reference "ANR-22-PECY-0011", within the framework of the "COMPROMIS" project).

\subsection*{Competing interests}
The authors have no relevant financial or non-financial interests to disclose.

\subsection*{Data availability}
The PASCAL VOC 2007 dataset \citep{VOC2007} used during the current study is available in the official PASCAL VOC repository: \url{https://www.robots.ox.ac.uk/~vgg/projects/pascal/VOC/voc2007/index.html}.

\subsection*{Code availability}
The source code for the metrics will be made publicly available upon publication.

\newpage

\begin{appendices}
\section{Comprehensive Attack Survey Data}
\label{app:survey_data}

The following tables provide the comprehensive data for our survey of adversarial attacks (Section \ref{sec:reviewATT}). As discussed in Section \ref{sec:criticsurvey}, this data highlights the fragmentation in the field. Table \ref{tab:AllATT} details the disparate outcomes, perturbation costs, metrics, detectors, and datasets reported in the original articles for a wide range of attacks.

\begin{sidewaystable} 
\scriptsize
\caption{Examples of adversarial attacks for object detection. * denotes if the code is publicly available. Inspired by \citet{BestSurvey}. }
\label{tab:AllATT}
\centering
\fontsize{6}{7}\selectfont
\begin{tabular*}{\textheight}{@{\extracolsep{\fill}} p{1.8cm} p{1.8cm} l l l p{2.2cm} p{3.5cm} p{2.5cm} @{}}
\toprule
NAME & OUTCOME & SPECIFICITY & LOCALITY & \makecell{PERTURBATION \\ COST} & METRICS & DETECTORS USED & DATASET USED \\
\midrule
AdvART \citep{AdvART} & Vanishing & Universal & Patch & SSIM & \makecell{mAP, \\ Success rate} &  \makecell{YOLOv4tiny, YOLOv3tiny, \\ YOLOv4, YOLOv3} &  \makecell{INRIA, MPII}\\
LGP \citep{LGP}* & Random & Image-specific & Entire image & \makecell{FID, IW-SSIM, \\ PSNR-B} & \makecell{mAP, Number \\of attacked targets per \\ image, Success rate} & \makecell{Faster R-CNN, Def-DETR, \\ VFNet, Cascade R-CNN, \\ Rotated FCOS,...} & \makecell{COCO, \\DOTA-v1.0}\\
OSFD \citep{OSFD}* & Random & Image-specific & Entire image & $L_\infty$ & mAP &  \makecell{Faster R-CNN, YOLOv3, \\ VFNet, Mask R-CNN \\ (Also tested on: DETR,\\ YOLOF, YOLOX, FCOS)} &  VOC2012\\
Patch-based FP \citep{PatchFP} & Fabrication & Universal & Patch & - & \makecell{mAP, \\ Average instances created, \\Average Precision Decrease, \\ Average Score Created, \\ FP Rate Increase} & \makecell{YOLOv3, YOLOv5s, \\ YOLOv5x} &  \makecell{DOTA-v1.0,\\RSOD, NWPU \\ VHR-10}\\
ShiftAttack \citep{ShiftAtt} & Other & Image-specific & Entire image & $L_2$, SSIM & Success rate &  \makecell{YOLOv3, YOLOv5, YOLOv8, \\ ATSS, FoveaBox, FreeAnchor, \\ GFL, RetinaNet, RTMDet, \\ Swin-Transformer, DDQ, \\ Def-DETR} &  \makecell{COCO, VOC2007, \\ VOC2012}\\
Targeted context attack \citep{TargConAtt} & Mislabeling & Image-specific & Entire image & $L_\infty$ & Success rate &  \makecell{Faster R-CNN, YOLOv3 \\ (Also tested on: Libra, \\ FoveaBox, FreeAnchor, DETR, \\Def-DETR} &  \makecell{COCO, VOC2007}\\
ASC \citep{ASC} & Vanishing & Image-specific & Entire image & $L_0$ & Success rate &  \makecell{SSD512, YOLOv3, YOLOX, \\ Faster R-CNN, Mask \\\ R-CNN, DETR, Def-DETR, \\ DAB-DETR} &  \makecell{COCO, \\ Cityscapes, \\ BDD100K}\\
EBAD \citep{EBAD}* & Mislabeling & Image-specific & Entire image & $L_\infty$ & Success rate &  \makecell{Faster R-CNN, YOLOv3, \\ FCOS, SSD, Grid R-CNN \\(Also tested on: RetinaNet, \\ Libra, Foveabox, FreeAnchor, \\ DETR, Def-DETR} &  \makecell{COCO, VOC2007}\\
Patches against UAV \citep{PatchesUAV}* & Vanishing & Universal  & Patch & -  & \makecell{mAP, \\ Success rate} &  \makecell{YOLOv5s, YOLOv3,\\YOLOv4} &  \makecell{VisDrone-2019}\\
Phantom Sponges \citep{PhantomSponges} & \makecell{Fabrication, \\ Inference latency} & Universal & Entire image & $L_2$ & \makecell{Number of created objects, \\ time, recall} &  \makecell{YOLOv3, YOLOv4, YOLOv5} &  \makecell{BDD100K, MTSD, \\ LISA, VOC} \\
T-SEA \citep{T-SEA}* & Vanishing & Universal & Patch & - & \makecell{AP \\ (for person)} & \makecell{Faster R-CNN, YOLOv2, \\ YOLOv3, YOLOv3-tiny, \\ YOLOv4, YOLOv4-tiny, \\ YOLOv5, SSD} &  \makecell{INRIA, \\ COCO-person, \\ CCTV-person}\\
ADC \citep{ADC} & \makecell{Mislabeling, Vanishing, \\ Fabrication} & Image-specific & Entire image & $L_\infty$ & Success rate &  \makecell{Faster R-CNN} &  \makecell{COCO,  VOC2007, \\ VOC2012}\\
\botrule
\multicolumn{8}{c}{\textit{Continued on next page}} \\
\botrule
\end{tabular*}
\end{sidewaystable} 

\begin{sidewaystable} 
\scriptsize
\caption*{\textit{Table \ref{tab:AllATT} - continued from previous page}}
\label{tab:AllATT2}
\centering
\fontsize{6}{7}\selectfont
\begin{tabular*}{\textwidth}{@{\extracolsep{\fill}} p{1.8cm} p{1.8cm} l l l p{2.2cm} p{3.5cm} p{2.5cm} @{}}
\toprule
NAME & OUTCOME & SPECIFICITY & LOCALITY & \makecell{PERTURBATION \\ COST} & METRICS & DETECTORS USED & DATASET USED \\
\midrule
\makecell{Adversarial Texture \\ \citep{AdvTexture}*} & \makecell{Vanishing} & Universal & Patch & - & AP &  \makecell{Faster R-CNN, YOLOv2, \\ YOLOv3, Mask R-CNN} &  \makecell{INRIA, \\ Custom dataset}\\
\makecell{CAA \citep{CAA}*} & Mislabeling & Image-specific & Entire image & $L_\infty$ & Success rate &  \makecell{Faster R-CNN, YOLOv3, \\ RetinaNet (Also tested on: \\ FCOS, Libra, Foveabox, \\ FreeAnchor, DETR, \\ Def-DETR} &  \makecell{COCO, VOC2007}\\
Daedalus \citep{Daedalus}* & Fabrication & Image-specific & Entire image & $L_0$, $L_2$ & mAP, FP rate &  \makecell{ YOLOv3, SSD, \\ RetinaNet} &  \makecell{COCO}\\
RAD \citep{RAD}* & Random & Image-specific & Entire image & $L_\infty$ & \makecell{mAP, Accuracy, \\ Mean Average Recall}&  \makecell{SSD, YOLOv3, RetinaNet, \\ Faster R-CNN, Mask \\ R-CNN, Cascade R-CNN, \\ Cascade Mask R-CNN, Hybrid \\ Task Cascade, EfficientDet} &  \makecell{COCO}\\
ZQA \citep{ZQA} & Mislabeling & Image-specific & Entire image & $L_\infty$ & Success rate &  \makecell{Faster R-CNN, Libra, \\ RetinaNet, FoveaBox} &  \makecell{COCO, VOC2007}\\
Naturalistic Patches \citep{NaturalisticPatches}* & Vanishing & Universal & Patch & - & \makecell{mAP} &  \makecell{YOLOv2, YOLOv3, YOLOv3-\\tiny, YOLOv4, YOLOv4-tiny,\\ Faster R-CNN} &  \makecell{INRIA, MPII, \\ Custom dataset}\\
Pick-Object Attack \citep{Pick-Object}* & Mislabeling & Image-specific & Entire image & $L_2$, SSIM & \makecell{mAP, Success rate, \\ Average Confidence \\ of True Class, \\ Average Confidence \\ of Adversarial Class} &  \makecell{Faster R-CNN} &  \makecell{COCO, \\ Visual Genome}\\
PRFA \citep{PRFA}* & Random & Image-specific & Patch & - & \makecell{maP, \\ Average queries} &  \makecell{Faster R-CNN, FCOS, \\ YOLOv3, ATSS} &  \makecell{COCO}\\
RPAttack \citep{RPAttack}* & Vanishing & Image-specific & Patch & - & AP & \makecell{Faster R-CNN, YOLOv4} &  \makecell{COCO, VOC2007}\\
U-DOS \citep{U-DOS} & Vanishing & Universal & Entire image & $L_\infty$ & \makecell{mAP, Blind degree} &  \makecell{Faster R-CNN, SSD, \\ YOLO} &  \makecell{VOC2007, VOC2012, \\ Caltech Pedestrian}\\
CAP \citep{CAP} & Vanishing & Image-specific & Entire image & \makecell{Peak Signal-to\\-Noise Ratio} & mAP, Recall &  \makecell{Faster R-CNN} &  \makecell{COCO, VOC2007}\\
Contextual Adversarial Patches \citep{ContAdvPatches}* & Vanishing & \makecell{Image-specific,\\Universal}  & Patch & -  & \makecell{mAP} &  \makecell{YOLOv2} &  \makecell{VOC2007, \\ KITTI}\\
\botrule
\end{tabular*}
\end{sidewaystable} 

\begin{sidewaystable} 
\scriptsize
\caption*{\textit{Table \ref{tab:AllATT} - continued from previous page}}
\centering
\fontsize{6}{7}\selectfont
\begin{tabular*}{\textwidth}{@{\extracolsep{\fill}} p{1.8cm} p{1.8cm} l l l p{2.2cm} p{3.5cm} p{2.5cm} @{}}
\toprule
NAME & OUTCOME & SPECIFICITY & LOCALITY & \makecell{PERTURBATION \\ COST} & METRICS & DETECTORS USED & DATASET USED \\
DPAttack \citep{DPAttack}* & Vanishing & Image-specific & Patch & - & AP & \makecell{Faster R-CNN, YOLOv4} &  COCO\\
Evaporate Attack \citep{Evaporate} & Vanishing & Image-specific & Entire image & $L_2$ & \makecell{Success rate, \\ False negative increase} &  \makecell{ YOLOv3, SSD \\ Faster R-CNN} &  \makecell{COCO, VOC2007}\\
\makecell{FB Invisible cloak \\ \citep{FbInvCloak}} & Vanishing & Universal & Patch & - & \makecell{AP} &  \makecell{YOLOv2, YOLOv3, \\ Faster R-CNN} &  \makecell{COCO, VOC2007, \\ INRIA, Custom \\ dataset}\\
TOG \citep{TOG}* & \makecell{Random, Vanishing, \\ Fabrication, Mislabeling} & \makecell{Image-specific, \\ Universal} & \makecell{Entire image, \\ Patch} & $L_\infty$ & \makecell{mAP, \% of \\ vanished objects} & \makecell{Faster R-CNN, SSD, \\ YOLOv3} &  \makecell{COCO, VOC2007, \\ INRIA} \\
UPC \citep{UPC}* & \makecell{Mislabeling} & \makecell{Universal} & \makecell{Patch \\ (on object)} & - & \makecell{AP} & \makecell{Faster R-CNN} &  \makecell{VOC2007, VOC2012, \\ Custom dataset} \\
\makecell{Adversarial T-shirt \\ \citep{AdvTshirt}*} & Vanishing & Universal & Patch & - & Success rate &  \makecell{Faster R-CNN, YOLOv2} &  \makecell{Custom dataset}\\
Patches to attack person detection \citep{PatchesPerson}* & Vanishing & Universal & Patch & - & Recall &  \makecell{YOLOv2} &  \makecell{INRIA}\\
Seeing isn't Believing \citep{SeeingisntBelieving} & Vanishing, Fabrication & Image-specific & \makecell{Patch} & - & \makecell{Success rate} &  \makecell{Faster R-CNN, YOLOv3, \\(Also tested on: SSD,\\ RFCN, Mask R-CNN} &  \makecell{Custom dataset}\\
ShapeShifter \citep{ShapeShifter}* & Mislabeling & Image-specific & \makecell{Patch \\ (on object)} & $L_\infty$ & \makecell{Success rate \\ (for stop sign)} &  \makecell{Faster R-CNN} &  \makecell{COCO}\\
DPatch \citep{DPatch}* & Random & Image-specific & Patch & - & mAP &  \makecell{Faster R-CNN, \\ YOLO} &  \makecell{VOC2007}\\
Imperceptible background patches \citep{ImpBackPatches} & Random & Image-specific & Patch & \makecell{Peak Signal-to\\-Noise Ratio} & \makecell{mAP, \\ True Positive Class \\ Loss, True Positive \\ Shape Loss, False \\ Positive Class Loss} &  \makecell{Faster R-CNN, SSD, \\ YOLOv3, YOLOv2} &  \makecell{COCO}\\
Invisible cloak \citep{InvCloak} & Vanishing & Universal & Patch & - & Precision &  \makecell{Tiny YOLO} &  \makecell{Custom dataset}\\
R-AP \citep{R-AP}* & Random & Image-specific & Entire image & \makecell{Peak Signal-to\\-Noise Ratio} & mAP &  \makecell{Faster R-CNN, Region \\ Fully Convolutional \\ Network} &  \makecell{COCO}\\
UEA \citep{UEA}* & Random & Image-specific & Entire image & - & mAP &  \makecell{Faster R-CNN, SSD} &  \makecell{VOC2007, \\ ImageNet VID}\\
DAG \citep{DAG}* & Mislabeling & Image-specific & Entire image & - & mAP, mIoU &  \makecell{Faster R-CNN, \\ FCN} &  \makecell{VOC2007, \\ Cityscapes}\\
\midrule
\botrule
\end{tabular*}
\end{sidewaystable} 

\section{Benchmark Perceptibility Metrics}
\label{app:perceptibility_data}

In Table \ref{tab:distancesbenchmark} we provide the detailed numerical data for the perceptibility analysis in Section \ref{sec:benchdist}. It lists the mean and standard deviation for $L_2$, $L_\infty$, SSIM, and LPIPS for each attack evaluated in our benchmark. 

\begin{sidewaystable}
\caption{Imperceptibility($L_2$,$L_\infty$,LPIPS)/Similarity(SSIM) of attacks 
between original and adversarial images ($\text{mean} \pm \text{std}$), depending on the surrogate model (Less perceptible in bold, second less perceptible underlined). For EBAD victim models are identical to surrogates. Note: For OSFD, $L_2$, LPIPS, and SSIM were computed after resizing images back to their original dimensions. The $L_\infty$ norm is reported at the attack's native resolution, as this is where its constraint is applied}
\label{tab:distancesbenchmark}
\centering
\renewcommand{\arraystretch}{1.4}
\small
\begin{tabular}{lccccc}
\toprule
\textbf{Attack} & \textbf{Surrogate Model} & \textbf{$L_2$↓} & \textbf{$L_\infty$↓} & \textbf{SSIM↑} & \textbf{LPIPS↓} \\
\midrule
\makecell[l]{CAA $\epsilon=10$} & F. R-CNN \& YOLOv3 & 4797.01 ± 379.45 & 43.53 ± 18.08 & \underline{0.84} ± 0.06 & \underline{0.13} ± 0.08 \\
\hline
\makecell[l]{CAA $\epsilon=30$} & F. R-CNN \& YOLOv3 & 9464.09 ± 717.02 & 56.11 ± 15.54 & 0.63 ± 0.11 & 0.31 ± 0.13 \\
\hline
\makecell[l]{EBAD $\epsilon=10$} & F. R-CNN \& YOLOv3 & \textbf{4223.71} ± 385.10 & \underline{10.0} ± 0.0 & \textbf{0.86} ± 0.06 & \textbf{0.11} ± 0.07 \\
\hline
\makecell[l]{EBAD $\epsilon=30$} & F. R-CNN \& YOLOv3 & 12339.27 ± 1169.83 & 30.0 ± 0.0 & 0.53 ± 0.13 & 0.40 ± 0.14 \\
\hline
\makecell[l]{EBAD $\epsilon=50$} & F. R-CNN \& YOLOv3 & 20281.73 ± 2065.19 & 50.0 ± 0.0 & 0.37 ± 0.12 & 0.62 ± 0.16 \\
\hline
\makecell[l]{OSFD} & YOLOv3 & 8255.96 ± 3090.36 & \textbf{7.0} ± 0.0 & 0.81 ± 0.04 & 0.20 ± 0.08 \\
\hline
\makecell[l]{OSFD} & F. R-CNN & 7707.84 ± 3080.61 & \textbf{7.0} ± 0.0 & 0.83 ± 0.05 & 0.17 ± 0.08 \\
\hline
\makecell[l]{Phantom Sponges} & YOLOv3 & 9745.94 ± 1258.09 & 89.96 ± 14.38 & 0.74 ± 0.08 & 0.23 ± 0.10 \\
\botrule
\end{tabular}
\end{sidewaystable}

\section{Detailed Adversarial Training Results (YOLOv3)}

The full experimental results for the YOLOv3 detector discussed in Section \ref{sec:defend} are presented in Table \ref{tab:mapyoloadvtrainannexe}. This table details the mAP performance of YOLOv3 models fine-tuned on various single-attack and mixed-attack datasets, evaluated against the complete suite of benchmark attacks.

\begin{sidewaystable} 
\caption{The \textbf{mAP}(\%) metric of the \textbf{YOLOv3} detectors after adversarial training. When only one attack and its percentage are given, the rest of the adversarial training dataset is filled with unattacked images. The number next to CAA and EBAD indicates the value of their respective parameters $\epsilon$. The Benign column concerns the original images. F. R-CNN is short for Faster R-CNN.}
\label{tab:mapyoloadvtrainannexe}
\centering
\fontsize{6}{7}\selectfont
\renewcommand{\arraystretch}{2} 
\begin{tabular}{ccccccccccccccc}
\toprule
\makecell{FINE-TUNED\\ON} & Benign & \makecell{CAA\\$\epsilon=10$} & \makecell{CAA\\$\epsilon=20$}& \makecell{CAA\\$\epsilon=30$} & \makecell{EBAD\\$\epsilon=10$} & \makecell{EBAD\\$\epsilon=30$} & \makecell{EBAD\\$\epsilon=50$} & \makecell{OSFD\\$k=1$\\YOLOv3} & \makecell{OSFD\\$k=2$\\YOLOv3} & \makecell{OSFD\\$k=3$\\YOLOv3} & \makecell{OSFD\\F. R-CNN} & \makecell{Phantom\\ Sponges \\YOLOv3}\\
\midrule
Baseline & \underline{75.8} & 12.3 & 7.3 & 6.0 & 22.0 & 19.3 & 18.1 & 20.2 & 8.1 & 6.6 & 22.4 & 63.7 \\
Benign & \underline{75.8} & 28.0 & 19.4 & 17.1 & 48.7 & 32.7 & 28.1 & 22.4 & 9.9 & 8.3 & 23.3 & 63.9 \\
$\text{CAA}_{30}$ & 72.2 & 72.2 & 69.3 & 66.3 & 72.1 & 67.4 & 60.6 & 59.8 & 54.0 & 53.0 & 61.6 & 66.8 \\
$\text{CAA}_{30}$ 75\% & 72.8 & 71.6 & 68.6 & 64.2 & 71.0 & 67.2 & 61.3 & 59.1 & 51.1 & 49.4 & 60.4 & 66.3 \\
$\text{CAA}_{30}$ 50\% & 73.6 & 69.5 & 65.5 & 61.1 & 69.8 & 65.1 & 59.4 & 53.1 & 43.6 & 41.4 & 54.9 & 66.1 \\
$\text{CAA}_{30}$ 25\% & 74.2 & 62.8 & 56.2 & 49.5 & 65.8 & 60.6 & 54.0 & 39.4 & 25.4 & 23.0 & 41.0 & 65.4 \\
$\text{EBAD}_{10}$ & 75.4 & 67.0 & 48.3 & 38.1 & 70.1 & 51.8 & 39.3 & 48.9 & 38.3 & 36.9 & 53.2 & 66.2 \\
$\text{EBAD}_{10}$ 75\% & 74.4 & 63.6 & 45.2 & 38.2 & 69.1 & 51.2 & 38.4 & 46.0 & 33.6 & 31.4 & 48.9 & 66.4 \\
$\text{EBAD}_{10}$ 50\% & 73.7 & 63.1 & 45.0 & 36.7 & 67.1 & 52.2 & 41.2 & 44.2 & 32.5 & 30.4 & 48.2 & 65.1 \\
$\text{EBAD}_{10}$ 25\% & 75.0 & 50.1 & 32.9 & 28.6 & 63.1 & 44.2 & 35.7 & 34.6 & 20.5 & 18.4 & 37.4 & 66.5 \\
OSFD & 73.5 & \textbf{73.9} & 69.6 & 65.1 & \textbf{73.6} & 65.2 & 53.5 & \textbf{71.1} & \textbf{69.7} & \textbf{69.1} & \textbf{69.8} & 65.4 \\
OSFD 75\% & 72.8 & 72.3 & 67.7 & 61.8 & 72.4 & 63.2 & 51.3 & 70.0 & 68.1 & 67.8 & 68.6 & 63.9 \\
OSFD 50\% & 71.8 & 71.2 & 66.4 & 60.5 & 71.6 & 62.5 & 51.1 & 69.0 & 66.5 & 66.0 & 67.1 & 62.5 \\
OSFD 25\% & 74.1 & 70.6 & 63.0 & 56.3 & 71.4 & 60.2 & 48.0 & 68.5 & 65.0 & 64.1 & 66.7 & 63.8 \\
OSFD 50\% $\text{CAA}_{30}$ 50\% & 72.2 & 72.9 & 70.9 & 67.6 & 72.8 & 67.6 & 59.5 & 69.2 & 67.6 & 67.2 & 68.5 & 66.8 \\
OSFD 75\% $\text{CAA}_{30}$ 25\% & 73.1 & 73.5 & 71.3 & 67.5 & \textbf{73.6} & 67.6 & 58.4 & \underline{70.7} & 68.9 & 68.2 & \underline{69.3} & 67.4 \\
\makecell{SFD 33\% $\text{CAA}_{30}$ 33\% \\ Benign 34\%} & 71.7 & 72.1 & 69.3 & 65.6 & 71.8 & 67.0 & 59.0 & 68.5 & 66.3 & 65.9 & 67.4 & 66.0 \\
PhantomSponges & 74.1 & 62.6 & 44.1 & 35.4 & 67.6 & 48.0 & 36.2 & 40.7 & 28.8 & 27.3 & 45.2 & \textbf{73.0} \\
PhantomSponges 75\% & 75.8 & 57.4 & 39.2 & 29.7 & 64.8 & 44.8 & 34.6 & 37.2 & 23.8 & 21.7 & 39.8 & \underline{72.6} \\
PhantomSponges 50\% & 75.5 & 52.1 & 34.7 & 25.3 & 60.7 & 42.0 & 32.4 & 32.6 & 20.0 & 17.9 & 35.2 & 71.8 \\
PhantomSponges 25\% & \textbf{75.9} & 37.4 & 24.9 & 22.3 & 56.4 & 36.5 & 30.6 & 26.7 & 13.5 & 11.9 & 28.8 & 70.1 \\
$\text{EBAD}_{50}$ & 68.7 & 69.4 & 68.9 & 66.8 & 70.0 & 68.8 & \underline{66.0} & 55.5 & 50.1 & 49.4 & 57.7 & 64.7 \\
$\text{EBAD}_{50}$ 50\% OSFD 50\% & 71.9 & 72.6 & \underline{71.6} & \underline{69.5} & 73.0 & 69.7 & 65.2 & 69.6 & 67.7 & 67.2 & 68.7 & 67.1 \\
$\text{EBAD}_{50}$ 75\% OSFD 25\% & 70.8 & 71.8 & 71.3 & \underline{69.5} & 71.9 & \underline{69.9} & \textbf{66.2} & 67.9 & 65.9 & 65.5 & 67.0 & 65.7 \\
$\text{EBAD}_{50}$ 25\% OSFD 75\% & 73.0 & \underline{73.7} & \textbf{71.9} & \textbf{69.6} & \underline{73.5} & \textbf{70.0} & 64.8 & \textbf{71.1} & \underline{69.3} & \underline{68.9} & \textbf{69.8} & 67.4 \\
\botrule
\end{tabular}
\end{sidewaystable} 

\section{Detailed Adversarial Training Results (Faster R-CNN)}

The full experimental results for the Faster R-CNN detector after adversarial training (cf. \ref{sec:defend}) are presented in Table \ref{tab:mapfrcnnadvtrainannexe}, followed by a granular breakdown of performance for the main experiments in Tables \ref{tab:mapfrcnnadvtrain}, \ref{tab:aplocfrcnnadvtrain}, and \ref{tab:csrfrcnnadvtrain}.

\begin{sidewaystable} 
\caption{The \textbf{mAP}(\%) metric of the \textbf{Faster R-CNN} detectors after adversarial training. When only one attack and its percentage are given, the rest of the adversarial dataset is filled with unattacked images. The number next to CAA and EBAD indicates the value of their respective parameters $\epsilon$. The Benign column concerns the original images. F. R-CNN is short for Faster R-CNN.}
\label{tab:mapfrcnnadvtrainannexe}
\centering
\fontsize{6}{7}\selectfont
\renewcommand{\arraystretch}{2} 
\begin{tabular}{ccccccccccc}
\toprule
\makecell{FINE-TUNED\\ON} & Benign & \makecell{CAA\\$\epsilon=10$} & \makecell{CAA\\$\epsilon=20$}& \makecell{CAA\\$\epsilon=30$} & \makecell{EBAD\\$\epsilon=10$\\F. R-CNN} & \makecell{EBAD\\$\epsilon=30$\\F. R-CNN} & \makecell{EBAD\\$\epsilon=50$\\F. R-CNN} & \makecell{OSFD\\$k=3$\\YOLOv3} & \makecell{OSFD\\F. R-CNN} & \makecell{Phantom\\ Sponges \\YOLOv3}\\
\midrule
Baseline & 79.3 & 8.7 & 3.9 & 3.4 & 28.9 & 26.1 & 24.9 & 7.4 & 3.4 & 65.7 \\
Benign & \textbf{80.4} & 19.8 & 9.3 & 8.0 & 38.0 & 31.8 & 30.2 & 6.9 & 3.6 & 65.1 \\
$\text{CAA}_{30}$ & 76.9 & \textbf{74.4} & \textbf{71.7} & \textbf{70.3} & 72.3 & 65.5 & 59.0 & 38.2 & 38.0 & 70.0 \\
$\text{CAA}_{30}$ 75\% & 77.8 & \underline{73.8} & 70.4 & 68.9 & 70.0 & 64.3 & 59.0 & 31.2 & 29.6 & 69.9 \\
$\text{CAA}_{30}$ 50\% & 78.8 & 72.8 & 69.1 & 67.4 & 66.2 & 63.4 & 59.2 & 23.1 & 20.4 & 70.0 \\
$\text{CAA}_{30}$ 25\% & \underline{79.6} & 70.3 & 66.5 & 64.5 & 59.8 & 60.4 & 56.9 & 16.9 & 12.9 & 69.7 \\
$\text{EBAD}_{10}$ & 76.7 & 73.7 & 69.5 & 67.5 & 72.6 & 63.7 & 55.7 & 35.2 & 35.4 & 69.4 \\
$\text{EBAD}_{10}$ 75\% & 77.4 & 72.6 & 68.5 & 66.4 & 71.9 & 64.2 & 57.4 & 32.5 & 32.6 & 69.0 \\
$\text{EBAD}_{10}$ 50\% & 78.1 & 72.1 & 67.1 & 64.7 & 70.6 & 63.7 & 57.6 & 27.3 & 27.4 & 69.4 \\
$\text{EBAD}_{10}$ 25\% & 79.0 & 70.3 & 63.9 & 60.8 & 67.3 & 61.0 & 53.1 & 21.1 & 19.6 & 68.9 \\
OSFD & 74.5 & 72.6 & 67.6 & 65.5 & 72.3 & 61.0 & 51.5 & \textbf{65.1} & \textbf{67.4} & 68.0 \\
OSFD 75\% & 75.7 & 72.4 & 68.2 & 65.9 & 71.9 & 62.6 & 54.1 & 63.7 & 65.8 & 68.2 \\
OSFD 50\% & 76.6 & 71.6 & 67.4 & 65.2 & 70.8 & 62.7 & 55.8 & 61.2 & 63.3 & 68.2 \\
OSFD 25\% & 77.8 & 71.0 & 65.6 & 63.4 & 69.0 & 61.9 & 55.7 & 56.6 & 58.0 & 68.2 \\
OSFD 50\% $\text{CAA}_{30}$ 50\% & 74.6 & 73.7 & \underline{71.4} & \underline{70.0} & \textbf{73.5} & 64.8 & 54.7 & 62.8 & 64.5 & 69.8 \\
OSFD 75\% $\text{CAA}_{30}$ 25\% & 74.5 & 73.5 & 70.5 & 69.1 & \underline{73.2} & 64.0 & 54.2 & \underline{64.4} & \underline{66.2} & 69.4 \\
\makecell{OSFD 33\% $\text{CAA}_{30}$ 33\% \\Benign 34\%} & 76.0 & 73.1 & 70.1 & 68.4 & 71.9 & 63.7 & 55.1 & 60.5 & 62.0 & 69.5 \\
PhantomSponges & 78.1 & 49.4 & 25.8 & 19.5 & 46.2 & 36.7 & 33.9 & 10.9 & 7.9 & \textbf{75.0} \\
PhantomSponges 75\% & 79.5 & 45.6 & 21.9 & 16.6 & 44.2 & 35.8 & 33.4 & 10.0 & 6.6 & \underline{74.6} \\
PhantomSponges 50\% & 80.0 & 41.1 & 18.9 & 14.7 & 43.2 & 35.5 & 33.2 & 9.6 & 5.9 & 73.9 \\
PhantomSponges 25\% & 80.3 & 33.3 & 14.9 & 12.1 & 41.2 & 34.5 & 32.6 & 8.5 & 4.8 & 72.8 \\
$\text{EBAD}_{50}$ & 74.9 & 72.0 & 69.5 & 68.6 & 72.2 & 68.4 & \underline{64.5} & 33.3 & 33.7 & 67.5 \\
$\text{EBAD}_{50}$ 50\% OSFD 50\% & 73.5 & 71.9 & 70.0 & 69.1 & 73.0 & \underline{68.7} & 63.7 & 61.7 & 63.6 & 68.1 \\
$\text{EBAD}_{50}$ 75\% OSFD 25\% & 72.4 & 70.9 & 69.2 & 68.7 & 72.3 & \textbf{68.9} & \textbf{64.8} & 58.8 & 60.6 & 67.0 \\
$\text{EBAD}_{50}$ 25\% OSFD 75\% & 74.3 & 72.5 & 70.0 & 68.8 & 73.1 & 67.5 & 62.1 & 63.7 & 65.8 & 68.6 \\
\botrule
\end{tabular}
\end{sidewaystable} 

\begin{sidewaystable}
\caption{Performance (\textbf{mAP} in \%) of \textbf{Faster R-CNN} detectors defended by adversarial training. The subscript next to CAA and EBAD indicates their respective $\epsilon$ parameters. For EBAD attacks, the target model is Faster R-CNN. For OSFD and Phantom Sponges, the model mentioned is the surrogate used for attack generation. Higher is better.}
\label{tab:mapfrcnnadvtrain}
\centering
\renewcommand{\arraystretch}{1.4} 
\begin{tabular}{c|ccccccccc} 
\toprule
\makecell{FINE-TUNED\\ON} & Benign & \makecell{$\text{CAA}_{10}$} & \makecell{$\text{CAA}_{30}$} & \makecell{$\text{EBAD}_{10}$} & \makecell{$\text{EBAD}_{30}$} & \makecell{$\text{EBAD}_{50}$} & \makecell{OSFD} & \makecell{OSFD} & \makecell{Phantom\\ Sponges}\\
& & & & & & & YOLOv3 & F. R-CNN & YOLOv3 \\
\cmidrule{2-10}
Baseline & \underline{79.3} & 8.7 & 3.4 & 28.9 & 26.1 & 24.9 & 7.4 & 3.4 & 65.7 \\
Benign & \textbf{80.4} & 19.8 & 8.0 & 38.0 & 31.8 & 30.2 & 6.9 & 3.6 & 65.1 \\
$\text{CAA}_{30}$ & 76.9 & \textbf{74.4} & \textbf{70.3} & \underline{72.3} & \underline{65.5} & \underline{59.0} & \underline{38.2} & \underline{38.0} & \underline{70.0} \\
$\text{EBAD}_{10}$ & 76.7 & \underline{73.7} & 67.5 & \textbf{72.6} & 63.7 & 55.7 & 35.2 & 35.4 & 69.4 \\
$\text{EBAD}_{50}$ & 74.9 & 72.0 & \underline{68.6} & 72.2 & \textbf{68.4} & \textbf{64.5} & 33.3 & 33.7 & 67.5 \\
OSFD & 74.5 & 72.6 & 65.5 & \underline{72.3} & 61.0 & 51.5 & \textbf{65.1} & \textbf{67.4} & 68.0 \\
\makecell{Phantom Sponges} & 78.1 & 49.4 & 19.5 & 46.2 & 36.7 & 33.9 & 10.9 & 7.9 & \textbf{75.0} \\
\botrule
\end{tabular}
\end{sidewaystable}

\begin{sidewaystable}
\caption{Performance (\textbf{AP\textsubscript{loc}} in \%) of \textbf{Faster R-CNN} detectors defended by adversarial training. The subscript next to CAA and EBAD indicates their respective $\epsilon$ parameters. For EBAD attacks, the target model is Faster R-CNN. For OSFD and Phantom Sponges, the model mentioned is the surrogate used for attack generation. Higher is better.}
\label{tab:aplocfrcnnadvtrain}
\centering
\renewcommand{\arraystretch}{1.4} 
\begin{tabular}{c|ccccccccc} 
\toprule
\makecell{FINE-TUNED\\ON} & Benign & \makecell{$\text{CAA}_{10}$} & \makecell{$\text{CAA}_{30}$} & \makecell{$\text{EBAD}_{10}$} & \makecell{$\text{EBAD}_{30}$} & \makecell{$\text{EBAD}_{50}$} & \makecell{OSFD} & \makecell{OSFD} & \makecell{Phantom\\ Sponges}\\
& & & & & & & YOLOv3 & F. R-CNN & YOLOv3 \\
\cmidrule{2-10}
Baseline & 82.7 & 37.3 & 33.3 & 67.4 & 66.1 & 65.4 & 4.0 & 2.1 & 73.3 \\
Benign & \textbf{84.7} & 45.3 & 36.1 & 69.3 & 68.2 & 67.7 & 4.7 & 2.5 & 73.1 \\
$\text{CAA}_{30}$ & 82.3 & \textbf{80.3} & \textbf{77.5} & 78.6 & \underline{74.8} & \underline{72.5} & \underline{45.9} & \underline{45.4} & \underline{76.5} \\
$\text{EBAD}_{10}$ & 82.4 & \underline{80.0} & 75.9 & \textbf{79.4} & \underline{74.8} & 72.1 & 42.3 & 43.1 & 76.0 \\
$\text{EBAD}_{50}$ & 81.1 & 79.0 & \underline{76.9} & \underline{79.0} & \textbf{76.7} & \textbf{74.9} & 40.9 & 41.0 & 74.9 \\
OSFD & 80.1 & 78.4 & 74.2 & 78.1 & 72.2 & 69.0 & \textbf{72.4} & \textbf{74.6} & 74.7 \\
\makecell{Phantom Sponges} & \underline{83.0} & 65.1 & 44.1 & 69.4 & 67.9 & 67.7 & 9.6 & 6.8 & \textbf{79.9} \\
\botrule
\end{tabular}
\end{sidewaystable}

\begin{sidewaystable}
\caption{Performance (\textbf{CSR} in \%) of \textbf{Faster R-CNN} detectors defended by adversarial training. The subscript next to CAA and EBAD indicates their respective $\epsilon$ parameters. For EBAD attacks, the target model is Faster R-CNN. For OSFD and Phantom Sponges, the model mentioned is the surrogate used for attack generation. Higher is better.}
\label{tab:csrfrcnnadvtrain}
\centering
\renewcommand{\arraystretch}{1.4} 
\begin{tabular}{c|ccccccccc} 
\toprule
\makecell{FINE-TUNED\\ON} & Benign & \makecell{$\text{CAA}_{10}$} & \makecell{$\text{CAA}_{30}$} & \makecell{$\text{EBAD}_{10}$} & \makecell{$\text{EBAD}_{30}$} & \makecell{$\text{EBAD}_{50}$} & \makecell{OSFD} & \makecell{OSFD} & \makecell{Phantom\\ Sponges}\\
& & & & & & & YOLOv3 & F. R-CNN & YOLOv3 \\
\cmidrule{2-10}
Baseline & \textbf{89.8} & 27.4 & 17.3 & 51.6 & 51.2 & 51.0 & 30.5 & 24.3 & 80.8 \\
Benign & \textbf{89.8} & 44.1 & 24.9 & 54.2 & 52.7 & 52.2 & 28.2 & 22.2 & 77.4 \\
$\text{CAA}_{30}$ & 88.5 & \underline{85.9} & \textbf{81.4} & 82.9 & \underline{77.6} & \underline{72.3} & \underline{65.8} & \underline{67.1} & \underline{82.5} \\
$\text{EBAD}_{10}$ & 88.7 & \textbf{86.0} & 79.7 & \textbf{83.9} & 76.6 & 70.2 & 64.0 & 66.5 & \underline{82.5} \\
$\text{EBAD}_{50}$ & 87.7 & 84.6 & \underline{81.1} & \underline{83.1} & \textbf{80.2} & \textbf{77.1} & 62.9 & 64.8 & 81.4 \\
OSFD & 86.6 & 84.0 & 77.2 & 82.8 & 73.4 & 66.0 & \textbf{80.5} & \textbf{82.0} & 80.1 \\
\makecell{Phantom Sponges} & \underline{89.1} & 67.5 & 45.1 & 61.9 & 55.2 & 53.8 & 39.4 & 37.5 & \textbf{85.5} \\
\botrule
\end{tabular}
\end{sidewaystable}

\end{appendices}

\newpage

\bibliography{sn-bibliography}

@inproceedings{OSFD,
  title={Transferable adversarial attacks for object detection using object-aware significant feature distortion},
  author={Ding, Xinlong and Chen, Jiansheng and Yu, Hongwei and Shang, Yu and Qin, Yining and Ma, Huimin},
  booktitle = {{Proceedings of the AAAI Conference on Artificial Intelligence}},
  year={2024}
}

@InProceedings{T-SEA,
    author    = {Huang, Hao and Chen, Ziyan and Chen, Huanran and Wang, Yongtao and Zhang, Kevin},
    title     = {{T-SEA}: {Transfer}-Based Self-Ensemble Attack on Object Detection},
    booktitle = {{Proceedings of the IEEE/CVF Conference on Computer Vision and Pattern Recognition (CVPR)}},
    year      = {2023}
}

@InProceedings{AdvtrainZhang,
author = {Zhang, Haichao and Wang, Jianyu},
title = {Towards Adversarially Robust Object Detection},
booktitle = {{Proceedings of the IEEE/CVF International Conference on Computer Vision (ICCV)}},
year = {2019}
}

@article{AdvtrainGabor,
  title={Adversarial defenses for object detectors based on {Gabor} convolutional layers},
  author={Amirkhani, Abdollah and Karimi, Mohammad Parsa},
  journal = {{The visual computer}},
  volume={38},
  number={6},
  pages={1929--1944},
  year={2022},
  publisher={Springer}
}

@InProceedings{AdvtrainChen,
    author    = {Chen, Pin-Chun and Kung, Bo-Han and Chen, Jun-Cheng},
    title     = {Class-Aware Robust Adversarial Training for Object Detection},
    booktitle = {{Proceedings of the IEEE/CVF Conference on Computer Vision and Pattern Recognition (CVPR)}},
    year      = {2021}
}

@article{LGP,
  title={Toward Generic and Controllable Attacks Against Object Detection},
  author={Li, Guopeng and Xu, Yue and Ding, Jian and Xia, Gui-Song},
  journal = {{IEEE Transactions on Geoscience and Remote Sensing}},
  volume={62},
  pages={1--12},
  year={2024},
  publisher={IEEE}
}

@article{ASC,
  title={To make yourself invisible with {Adversarial Semantic Contours}},
  author={Zhang, Yichi and Zhu, Zijian and Su, Hang and Zhu, Jun and Zheng, Shibao and He, Yuan and Xue, Hui},
  journal = {{Computer Vision and Image Understanding}},
  volume={230},
  pages={103659},
  year={2023},
  publisher={Elsevier}
}

@article{TargConAtt,
  title={Targeted context attack for object detection},
  author={Sun, Changfeng and Zhang, Xuchong and Han, Haoliang and Sun, Hongbin},
  journal = {{Neurocomputing}},
  volume={601},
  pages={128208},
  year={2024},
  publisher={Elsevier}
}

@article{ShiftAtt,
  title={{ShiftAttack}: {Toward} Attacking the Localization Ability of Object Detector},
  author={Li, Hao and Yang, Zeyu and Gong, Maoguo and Chen, Shiguo and Qin, AK and Niu, Zhenxing and Wu, Yue and Zhou, Yu},
  journal = {{IEEE Transactions on Circuits and Systems for Video Technology}},
  year={2024},
  publisher={IEEE}
}

@InProceedings{PhantomSponges,
    author    = {Shapira, Avishag and Zolfi, Alon and Demetrio, Luca and Biggio, Battista and Shabtai, Asaf},
    title     = {{Phantom Sponges}: Exploiting Non-Maximum Suppression To Attack Deep Object Detectors},
    booktitle = {{Proceedings of the IEEE/CVF Winter Conference on Applications of Computer Vision (WACV)}},
    year      = {2023}
}

@InProceedings{ADC,
    author    = {Yin, Mingjun and Li, Shasha and Song, Chengyu and Asif, M. Salman and Roy-Chowdhury, Amit K. and Krishnamurthy, Srikanth V.},
    title     = {{ADC}: {Adversarial} Attacks Against Object Detection That Evade Context Consistency Checks},
    booktitle = {{Proceedings of the IEEE/CVF Winter Conference on Applications of Computer Vision (WACV)}},
    year      = {2022}
}

@InProceedings{EBAD,
    author    = {Cai, Zikui and Tan, Yaoteng and Asif, M. Salman},
    title     = {Ensemble-Based Blackbox Attacks on Dense Prediction},
    booktitle = {{Proceedings of the IEEE/CVF Conference on Computer Vision and Pattern Recognition (CVPR)}},
    year      = {2023}
}

@inproceedings{CAA,
  title={Context-aware transfer attacks for object detection},
  author={Cai, Zikui and Xie, Xinxin and Li, Shasha and Yin, Mingjun and Song, Chengyu and Krishnamurthy, Srikanth V and Roy-Chowdhury, Amit K and Asif, M Salman},
  booktitle = {{Proceedings of the AAAI Conference on Artificial Intelligence}},
  year={2022}
}

@inproceedings{TOG,
  title={Adversarial objectness gradient attacks in real-time object detection systems},
  author={Chow, Ka-Ho and Liu, Ling and Loper, Margaret and Bae, Juhyun and Gursoy, Mehmet Emre and Truex, Stacey and Wei, Wenqi and Wu, Yanzhao},
  booktitle = {{2020 Second IEEE International Conference on Trust, Privacy and Security in Intelligent Systems and Applications (TPS-ISA)}},
  year={2020}
}

@inproceedings{RPAttack,
  author={Huang, Hao and Wang, Yongtao and Chen, Zhaoyu and Tang, Zhi and Zhang, Wenqiang and Ma, Kai-Kuang},
  booktitle = {{2021 IEEE International Conference on Multimedia and Expo (ICME)}}, 
  title={{RPAttack}: Refined Patch Attack on General Object Detectors}, 
  year={2021}
}

@article{DPAttack,
  title={{DPAttack}: {Diffused} Patch Attacks against Universal Object Detection},
  author={Wu, Shudeng and Dai, Tao and Xia, Shu-Tao},
  journal = {{arXiv preprint arXiv:2010.11679}},
  year={2020}
}

@InProceedings{ZQA,
    author    = {Cai, Zikui and Rane, Shantanu and Brito, Alejandro E. and Song, Chengyu and Krishnamurthy, Srikanth V. and Roy-Chowdhury, Amit K. and Asif, M. Salman},
    title     = {Zero-Query Transfer Attacks on Context-Aware Object Detectors},
    booktitle = {{Proceedings of the IEEE/CVF Conference on Computer Vision and Pattern Recognition (CVPR)}},
    year      = {2022}
}

@article{PatchFP,
  title={Adversarial patch-based false positive creation attacks against aerial imagery object detectors},
  author={Tang, Guijian and Yao, Wen and Jiang, Tingsong and Zhao, Yong and Sun, Jialiang},
  journal = {{Neurocomputing}},
  volume={579},
  pages={127431},
  year={2024},
  publisher={Elsevier}
}

@article{Daedalus,
  title={Daedalus: {Breaking} Nonmaximum Suppression in Object Detection via Adversarial Examples},
  author={Wang, Derui and Li, Chaoran and Wen, Sheng and Han, Qing-Long and Nepal, Surya and Zhang, Xiangyu and Xiang, Yang},
  journal = {{IEEE Transactions on Cybernetics}},
  volume={52},
  number={8},
  pages={7427--7440},
  year={2021},
  publisher={IEEE}
}

@article{Evaporate,
  title={An adversarial attack on {DNN-based} black-box object detectors},
  author={Wang, Yajie and Tan, Yu-an and Zhang, Wenjiao and Zhao, Yuhang and Kuang, Xiaohui},
  journal = {{Journal of Network and Computer Applications}},
  volume={161},
  pages={102634},
  year={2020},
  publisher={Elsevier}
}

@article{Pick-Object,
  title = {{Pick-Object-Attack}: Type-specific adversarial attack for object detection},
  author={Nezami, Omid Mohamad and Chaturvedi, Akshay and Dras, Mark and Garain, Utpal},
  journal = {{Computer Vision and Image Understanding}},
  volume={211},
  pages={103257},
  year={2021},
  publisher={Elsevier}
}

@article{RAD,
  title={Relevance attack on detectors},
  author={Chen, Sizhe and He, Fan and Huang, Xiaolin and Zhang, Kun},
  journal = {{Pattern Recognition}},
  volume={124},
  pages={108491},
  year={2022},
  publisher={Elsevier}
}

@inproceedings{CAP,
  title={Contextual adversarial attacks for object detection},
  author={Zhang, Hantao and Zhou, Wengang and Li, Houqiang},
  booktitle = {{2020 IEEE international conference on multimedia and expo (ICME)}},
  year={2020}
}

@inproceedings{UEA,
  title     = {Transferable Adversarial Attacks for Image and Video Object Detection},
  author    = {Wei, Xingxing and Liang, Siyuan and Chen, Ning and Cao, Xiaochun},
  booktitle = {{Proceedings of the Twenty-Eighth International Joint Conference on Artificial Intelligence (IJCAI)}},
  year      = {2019}
}

@inproceedings{R-AP,
  author={Li, Yuezun and Tian, Daniel and Chang, Ming-Ching and Bian, Xiao and Lyu, Siwei},
  title        = {Robust Adversarial Perturbation on Deep Proposal-based Models},
  booktitle    = {British Machine Vision Conference 2018, {BMVC}},
  year         = {2018}
}

@InProceedings{DAG,
author = {Xie, Cihang and Wang, Jianyu and Zhang, Zhishuai and Zhou, Yuyin and Xie, Lingxi and Yuille, Alan},
title = {Adversarial Examples for Semantic Segmentation and Object Detection},
booktitle = {{Proceedings of the IEEE International Conference on Computer Vision (ICCV)}},
year = {2017}
}

@article{DPatch,
  title={{DPatch}: {Attacking} Object Detectors with Adversarial Patches},
  author={Liu, Xin and Yang, Huanrui and Song, Linghao and Li, Hai and Chen, Yiran},
  journal = {{CoRR, abs/1806.02299}},
  volume={2},
  pages={1},
  year={2018}
}

@InProceedings{PRFA,
    author    = {Liang, Siyuan and Wu, Baoyuan and Fan, Yanbo and Wei, Xingxing and Cao, Xiaochun},
    title     = {Parallel Rectangle Flip Attack: {A} Query-Based Black-Box Attack Against Object Detection},
    booktitle = {{Proceedings of the IEEE/CVF International Conference on Computer Vision (ICCV)}},
    year      = {2021}
}

@inproceedings{AdvART,
  title={{AdvART}: {Adversarial} art for camouflaged object detection attacks},
  author={Guesmi, Amira and Bilasco, Ioan Marius and Shafique, Muhammad and Alouani, Ihsen},
  booktitle = {{2024 IEEE International Conference on Image Processing (ICIP)}},
  year={2024}
}

@inproceedings{InvCloak,
  title={Building Towards "{Invisible Cloak}": {Robust} Physical Adversarial Attack on {YOLO} Object Detector},
  author={Yang, Darren Yu and Xiong, Jay and Li, Xincheng and Yan, Xu and Raiti, John and Wang, Yuntao and Wu, HuaQiang and Zhong, Zhenyu},
  booktitle = {{2018 9th IEEE Annual Ubiquitous Computing, Electronics \& Mobile Communication Conference (UEMCON)}},
  year={2018}
}

@article{ImpBackPatches,
  title={Attacking object detectors via imperceptible patches on background},
  author={Li, Yuezun and Bian, Xian and Lyu, Siwei},
  journal = {{arXiv preprint arXiv:1809.05966}},
  volume={1},
  year={2018}
}

@InProceedings{PatchesPerson,
author = {Thys, Simen and Van Ranst, Wiebe and Goedeme, Toon},
title = {Fooling Automated Surveillance Cameras: {Adversarial} Patches to Attack Person Detection},
booktitle = {{Proceedings of the IEEE/CVF Conference on Computer Vision and Pattern Recognition (CVPR) Workshops}},
year = {2019}
}

@article{U-DOS,
  title={Universal adversarial perturbations against object detection},
  author={Li, Debang and Zhang, Junge and Huang, Kaiqi},
  journal = {{Pattern Recognition}},
  volume={110},
  pages={107584},
  year={2021},
  publisher={Elsevier}
}

@inproceedings{ShapeShifter,
  title={{ShapeShifter}: {Robust} Physical Adversarial Attack on {Faster R-CNN} Object Detector},
  author={Chen, Shang-Tse and Cornelius, Cory and Martin, Jason and Chau, Duen Horng},
  booktitle = {{Joint European Conference on Machine Learning and Knowledge Discovery in Databases (ECML PKDD)}},
  year={2018}
}

@INPROCEEDINGS{PatchesUAV,
  author={Shrestha, Samridha and Pathak, Saurabh and Viegas, Eduardo K.},
  booktitle = {{2023 IEEE/RSJ International Conference on Intelligent Robots and Systems (IROS)}}, 
  title={Towards a Robust Adversarial Patch Attack Against Unmanned Aerial Vehicles Object Detection}, 
  year={2023}
}

@article{FirstSurvey,
  title={A survey on adversarial attacks and defenses for object detection and their applications in autonomous vehicles},
  author={Amirkhani, Abdollah and Karimi, Mohammad Parsa and Banitalebi-Dehkordi, Amin},
  journal = {{The Visual Computer}},
  volume={39},
  number={11},
  pages={5293--5307},
  year={2023},
  publisher={Springer}
}

@ARTICLE{BestSurvey,
  author={Nguyen, Khoi Nguyen Tiet and Zhang, Wenyu and Lu, Kangkang and Wu, Yu-Huan and Zheng, Xingjian and Li Tan, Hui and Zhen, Liangli},
  journal = {{IEEE Transactions on Neural Networks and Learning Systems}}, 
  title={A Survey and Evaluation of Adversarial Attacks in Object Detection}, 
  year={2025},
  volume={36},
  number={9},
  pages={15706-15722},
}

@InProceedings{ContAdvPatches,
author = {Saha, Aniruddha and Subramanya, Akshayvarun and Patil, Koninika and Pirsiavash, Hamed},
title = {Role of Spatial Context in Adversarial Robustness for Object Detection},
booktitle = {{Proceedings of the IEEE/CVF Conference on Computer Vision and Pattern Recognition (CVPR) Workshops}},
year = {2020}
}

@InProceedings{SAC,
    author    = {Liu, Jiang and Levine, Alexander and Lau, Chun Pong and Chellappa, Rama and Feizi, Soheil},
    title     = {{Segment and Complete}: {Defending Object} Detectors Against Adversarial Patch Attacks With Robust Patch Detection},
    booktitle = {{Proceedings of the IEEE/CVF Conference on Computer Vision and Pattern Recognition (CVPR)}},
    year      = {2022}
}

@inproceedings{AdvPatchFeatEnergy,
  title={Defending physical adversarial attack on object detection via adversarial patch-feature energy},
  author={Kim, Taeheon and Yu, Youngjoon and Ro, Yong Man},
  booktitle = {{Proceedings of the 30th ACM International Conference on Multimedia}},
  year={2022}
}

@inproceedings{SeeingisntBelieving,
  author    = {Yue Zhao and Hong Zhu and Ruigang Liang and Qintao Shen and Shengzhi Zhang and Kai Chen},
  title     = {{Seeing isn't Believing}: {Towards} More Robust Adversarial Attack Against Real World Object Detectors},
  booktitle = {{Proceedings of the 2019 ACM SIGSAC Conference on Computer and Communications Security}},
  year      = {2019}
}

@inproceedings{ContextInconsistency,
  title={Connecting the dots: {Detecting} adversarial perturbations using context inconsistency},
  author={Li, Shasha and Zhu, Shitong and Paul, Sudipta and Roy-Chowdhury, Amit and Song, Chengyu and Krishnamurthy, Srikanth and Swami, Ananthram and Chan, Kevin S},
  booktitle = {{European Conference on Computer Vision (ECCV)}},
  year={2020}
}

@article{ClassifSurvey,
  title={How Deep Learning Sees the World: {A} Survey on Adversarial Attacks \& Defenses},
  author={Costa, Joana C and Roxo, Tiago and Proen{\c{c}}a, Hugo and Inacio, Pedro Ricardo Morais},
  journal = {{IEEE Access}},
  volume={12},
  pages={61113--61136},
  year={2024},
  publisher={IEEE}
}

@article{ClassifSurveyCAAI,
  title={A survey on adversarial attacks and defences},
  author={Chakraborty, Anirban and Alam, Manaar and Dey, Vishal and Chattopadhyay, Anupam and Mukhopadhyay, Debdeep},
  journal = {{CAAI Transactions on Intelligence Technology}},
  volume={6},
  number={1},
  pages={25--45},
  year={2021},
  publisher={Wiley Online Library}
}

@article{ClassifSurveyDefense,
  title={Adversarial learning targeting deep neural network classification: A comprehensive review of defenses against attacks},
  author={Miller, David J and Xiang, Zhen and Kesidis, George},
  journal = {{Proceedings of the IEEE}},
  volume={108},
  number={3},
  pages={402--433},
  year={2020},
  publisher={IEEE}
}

@InProceedings{ReID,
author = {Bouniot, Quentin and Audigier, Romaric and Loesch, Angelique},
title = {Vulnerability of Person Re-Identification Models to Metric Adversarial Attacks},
booktitle = {{Proceedings of the IEEE/CVF Conference on Computer Vision and Pattern Recognition (CVPR) Workshops}},
year = {2020}
}

@InProceedings{UPC,
author = {Huang, Lifeng and Gao, Chengying and Zhou, Yuyin and Xie, Cihang and Yuille, Alan L. and Zou, Changqing and Liu, Ning},
title = {Universal Physical Camouflage Attacks on Object Detectors},
booktitle = {{Proceedings of the IEEE/CVF Conference on Computer Vision and Pattern Recognition (CVPR)}},
year = {2020}
}

@inproceedings{AdvTshirt,
  title={Adversarial {T-shirt}! {Evading} person detectors in a physical world},
  author={Xu, Kaidi and Zhang, Gaoyuan and Liu, Sijia and Fan, Quanfu and Sun, Mengshu and Chen, Hongge and Chen, Pin-Yu and Wang, Yanzhi and Lin, Xue},
  booktitle = {{European Conference on Computer Vision (ECCV)}},
  year={2020}
}

@InProceedings{NaturalisticPatches,
    author    = {Hu, Yu-Chih-Tuan and Kung, Bo-Han and Tan, Daniel Stanley and Chen, Jun-Cheng and Hua, Kai-Lung and Cheng, Wen-Huang},
    title     = {Naturalistic Physical Adversarial Patch for Object Detectors},
    booktitle = {{Proceedings of the IEEE/CVF International Conference on Computer Vision (ICCV)}},
    year      = {2021}
}

@InProceedings{AdvTexture,
    author    = {Hu, Zhanhao and Huang, Siyuan and Zhu, Xiaopei and Sun, Fuchun and Zhang, Bo and Hu, Xiaolin},
    title     = {Adversarial Texture for Fooling Person Detectors in the Physical World},
    booktitle = {{Proceedings of the IEEE/CVF Conference on Computer Vision and Pattern Recognition (CVPR)}},
    year      = {2022}
}

@inproceedings{FbInvCloak,
  title={Making an Invisibility Cloak: {Real} World Adversarial Attacks on Object Detectors},
  author={Wu, Zuxuan and Lim, Ser-Nam and Davis, Larry S and Goldstein, Tom},
  booktitle = {{European Conference on Computer Vision (ECCV)}},
  year={2020}
}

@inproceedings{LPIPS,
  title={The unreasonable effectiveness of deep features as a perceptual metric},
  author={Zhang, Richard and Isola, Phillip and Efros, Alexei A and Shechtman, Eli and Wang, Oliver},
  booktitle = {{Proceedings of the IEEE/CVF Conference on Computer Vision and Pattern Recognition (CVPR)}},
  year={2018}
}

@article{mmdetection,
  title   = {{MMDetection}: {Open MMLab} Detection Toolbox and Benchmark},
  author  = {Chen, Kai and Wang, Jiaqi and Pang, Jiangmiao and Cao, Yuhang and
             Xiong, Yu and Li, Xiaoxiao and Sun, Shuyang and Feng, Wansen and
             Liu, Ziwei and Xu, Jiarui and Zhang, Zheng and Cheng, Dazhi and
             Zhu, Chenchen and Cheng, Tianheng and Zhao, Qijie and Li, Buyu and
             Lu, Xin and Zhu, Rui and Wu, Yue and Dai, Jifeng and Wang, Jingdong
             and Shi, Jianping and Ouyang, Wanli and Loy, Chen Change and Lin, Dahua},
  journal = {{arXiv preprint arXiv:1906.07155}},
  year={2019}
}

@inproceedings{COCO,
  title={Microsoft {COCO}: {Common} Objects in Context},
  author={Lin, Tsung-Yi and Maire, Michael and Belongie, Serge and Hays, James and Perona, Pietro and Ramanan, Deva and Doll{\'a}r, Piotr and Zitnick, C Lawrence},
  booktitle = {{European Conference on Computer Vision (ECCV)}},
  year={2014}
}

@article{VOC2007,
  title={The {Pascal Visual Object Classes} ({VOC}) Challenge},
  author={Everingham, Mark and Van Gool, Luc and Williams, Christopher KI and Winn, John and Zisserman, Andrew},
  journal = {{International journal of computer vision}},
  volume={88},
  number={2},
  pages={303--338},
  year={2010},
  publisher={Springer}
}

@article{yolov3,
  title={Yolov3: An incremental improvement},
  author={Redmon, Joseph and Farhadi, Ali},
  journal = {{arXiv preprint arXiv:1804.02767}},
  year={2018}
}

@InProceedings{resnet,
author = {He, Kaiming and Zhang, Xiangyu and Ren, Shaoqing and Sun, Jian},
title = {Deep Residual Learning for Image Recognition},
booktitle = {{Proceedings of the IEEE/CVF Conference on Computer Vision and Pattern Recognition (CVPR)}},
year = {2016}
}

@inproceedings{DETR,
  title={End-to-end object detection with transformers},
  author={Carion, Nicolas and Massa, Francisco and Synnaeve, Gabriel and Usunier, Nicolas and Kirillov, Alexander and Zagoruyko, Sergey},
  booktitle = {{European Conference on Computer Vision (ECCV)}},
  year={2020}
}

@InProceedings{bdd100k,
author = {Yu, Fisher and Chen, Haofeng and Wang, Xin and Xian, Wenqi and Chen, Yingying and Liu, Fangchen and Madhavan, Vashisht and Darrell, Trevor},
title = {{BDD100K}: A Diverse Driving Dataset for Heterogeneous Multitask Learning},
booktitle = {{Proceedings of the IEEE/CVF Conference on Computer Vision and Pattern Recognition (CVPR)}},
year = {2020}
}

@inproceedings{TRADES,
  title={Theoretically principled trade-off between robustness and accuracy},
  author={Zhang, Hongyang and Yu, Yaodong and Jiao, Jiantao and Xing, Eric and El Ghaoui, Laurent and Jordan, Michael},
  booktitle = {{International conference on machine learning (ICML)}},
  year={2019}
}

@inproceedings{goodfellow,
  author       = {Ian J. Goodfellow and
                  Jonathon Shlens and
                  Christian Szegedy},
  editor       = {Yoshua Bengio and
                  Yann LeCun},
  title        = {Explaining and Harnessing Adversarial Examples},
  booktitle = {{3rd International Conference on Learning Representations, (ICLR)}},
  year         = {2015}
}

@article{Yolov4,
  title={{YOLOv4}: {Optimal} Speed and Accuracy of Object Detection},
  author={Bochkovskiy, Alexey and Wang, Chien-Yao and Liao, Hong-Yuan Mark},
  journal = {{arXiv preprint arXiv:2004.10934}},
  year={2020}
}

@inproceedings{SSD,
  title={{{SSD}}: {{Single}} Shot {{MultiBox}} Detector},
  author={Liu, Wei and Anguelov, Dragomir and Erhan, Dumitru and Szegedy, Christian and Reed, Scott and Fu, Cheng-Yang and Berg, Alexander C},
  booktitle = {{European Conference on Computer Vision (ECCV)}},
  pages={21--37},
  year={2016},
  organization={Springer}
}

@InProceedings{RetinaNet,
author = {Lin, Tsung-Yi and Goyal, Priya and Girshick, Ross and He, Kaiming and Dollar, Piotr},
title = {Focal Loss for Dense Object Detection},
booktitle = {{Proceedings of the IEEE International Conference on Computer Vision (ICCV)}},
year = {2017}
}

@InProceedings{Fcos,
author = {Tian, Zhi and Shen, Chunhua and Chen, Hao and He, Tong},
title = {{FCOS}: {Fully} Convolutional One-Stage Object Detection},
booktitle = {{Proceedings of the IEEE/CVF International Conference on Computer Vision (ICCV)}},
year = {2019}
}

@InProceedings{CenterNet,
author = {Duan, Kaiwen and Bai, Song and Xie, Lingxi and Qi, Honggang and Huang, Qingming and Tian, Qi},
title = {{CenterNet}: {Keypoint} Triplets for Object Detection},
booktitle = {{Proceedings of the IEEE/CVF International Conference on Computer Vision (ICCV)}},
year = {2019}
}

@article{YOLOX,
  title={Yolox: {Exceeding} yolo series in 2021},
  author={Ge, Zheng and Liu, Songtao and Wang, Feng and Li, Zeming and Sun, Jian},
  journal = {{arXiv preprint arXiv:2107.08430}},
  year={2021}
}

@inproceedings{
DeformableDETR,
title={Deformable {DETR}: Deformable Transformers for End-to-End Object Detection},
author={Xizhou Zhu and Weijie Su and Lewei Lu and Bin Li and Xiaogang Wang and Jifeng Dai},
booktitle = {{International Conference on Learning Representations (ICLR)}},
year={2021}
}

@article{faster-r-cnn,
  title={{Faster R-CNN}: {Towards} Real-Time Object Detection with Region Proposal Networks},
  author={Ren, Shaoqing and He, Kaiming and Girshick, Ross and Sun, Jian},
  journal = {{Advances in neural information processing systems (NeurIPS)}},
  year={2015}
}

@InProceedings{Cascader-cnn,
author = {Cai, Zhaowei and Vasconcelos, Nuno},
title = {{Cascade R-CNN}: {Delving} Into High Quality Object Detection},
booktitle = {{Proceedings of the IEEE/CVF Conference on Computer Vision and Pattern Recognition (CVPR)}},
year = {2018}
}

@InProceedings{Librar-cnn,
author = {Pang, Jiangmiao and Chen, Kai and Shi, Jianping and Feng, Huajun and Ouyang, Wanli and Lin, Dahua},
title = {{Libra R-CNN}: {Towards} Balanced Learning for Object Detection},
booktitle = {{Proceedings of the IEEE/CVF Conference on Computer Vision and Pattern Recognition (CVPR)}},
year = {2019}
}

@InProceedings{Sparser-cnn,
    author    = {Sun, Peize and Zhang, Rufeng and Jiang, Yi and Kong, Tao and Xu, Chenfeng and Zhan, Wei and Tomizuka, Masayoshi and Li, Lei and Yuan, Zehuan and Wang, Changhu and Luo, Ping},
    title     = {{Sparse R-CNN}: {End}-to-End Object Detection With Learnable Proposals},
    booktitle = {{Proceedings of the IEEE/CVF Conference on Computer Vision and Pattern Recognition (CVPR)}},
    year      = {2021}
}

@InProceedings{Querydet,
    author    = {Yang, Chenhongyi and Huang, Zehao and Wang, Naiyan},
    title     = {{QueryDet}: {Cascaded} Sparse Query for Accelerating High-Resolution Small Object Detection},
    booktitle = {{Proceedings of the IEEE/CVF Conference on Computer Vision and Pattern Recognition (CVPR)}},
    year      = {2022}
}

@inproceedings{RobustDet,
  title={Adversarially-aware robust object detector},
  author={Dong, Ziyi and Wei, Pengxu and Lin, Liang},
  booktitle = {{European Conference on Computer Vision (ECCV)}},
  year={2022}
}

@article{AdvRecent,
  title={On the importance of backbone to the adversarial robustness of object detectors},
  author={Li, Xiao and Chen, Hang and Hu, Xiaolin},
  journal = {{IEEE Transactions on Information Forensics and Security}},
  year={2025},
  publisher={IEEE}
}

@article{AdvMeta,
  title={Meta adversarial training against universal patches},
  author={Metzen, Jan Hendrik and Finnie, Nicole and Hutmacher, Robin},
  journal = {{arXiv preprint arXiv:2101.11453}},
  year={2021}
}

@inproceedings{filter,
  title={Preprocessing-based adversarial defense for object detection via feature filtration},
  author={Zhou, Ling and Liu, Qihe and Zhou, Shijie},
  booktitle = {{Proceedings of the 7th International Conference on Algorithms, Computing and Systems}},
  year={2023}
}

@InProceedings{JEDI,
    author    = {Tarchoun, Bilel and Ben Khalifa, Anouar and Mahjoub, Mohamed Ali and Abu-Ghazaleh, Nael and Alouani, Ihsen},
    title     = {Jedi: {Entropy}-Based Localization and Removal of Adversarial Patches},
    booktitle = {{Proceedings of the IEEE/CVF Conference on Computer Vision and Pattern Recognition (CVPR)}},
    year      = {2023}
}

@InProceedings{PAD,
    author    = {Jing, Lihua and Wang, Rui and Ren, Wenqi and Dong, Xin and Zou, Cong},
    title     = {{PAD}: {Patch}-Agnostic Defense against Adversarial Patch Attacks},
    booktitle = {{Proceedings of the IEEE/CVF Conference on Computer Vision and Pattern Recognition (CVPR)}},
    year      = {2024}
}

@article{CertDet,
  title={Detection as regression: {Certified} object detection with median smoothing},
  author={Chiang, Ping-yeh and Curry, Michael and Abdelkader, Ahmed and Kumar, Aounon and Dickerson, John and Goldstein, Tom},
  journal = {{Advances in Neural Information Processing Systems (NeurIPS)}},
  year={2020}
}

@inproceedings{detectorguard,
  title={{DetectorGuard}: {Provably} securing object detectors against localized patch hiding attacks},
  author={Xiang, Chong and Mittal, Prateek},
  booktitle = {{Proceedings of the 2021 ACM SIGSAC conference on computer and communications security}},
  year={2021}
}

@inproceedings{objectseeker,
  title={{ObjectSeeker}: {Certifiably} robust object detection against patch hiding attacks via patch-agnostic masking},
  author={Xiang, Chong and Valtchanov, Alexander and Mahloujifar, Saeed and Mittal, Prateek},
  booktitle = {{2023 IEEE Symposium on Security and Privacy (SP)}},
  year={2023}
}

@inproceedings{TIDE,
  title={{TIDE}: {A} general toolbox for identifying object detection errors},
  author={Bolya, Daniel and Foley, Sean and Hays, James and Hoffman, Judy},
  booktitle = {{European Conference on Computer Vision (ECCV)}},
  year={2020}
}

@inproceedings{diagnosing,
  title={Diagnosing error in object detectors},
  author={Hoiem, Derek and Chodpathumwan, Yodsawalai and Dai, Qieyun},
  booktitle = {{European Conference on Computer Vision }},
  year={2012}
}

@inproceedings{AdvDiff,
  title={Diffusion Models for Adversarial Purification},
  author={Nie, Weili and Guo, Brandon and Huang, Yujia and Xiao, Chaowei and Vahdat, Arash and Anandkumar, Animashree},
  booktitle = {{Proceedings of the 39th International Conference on Machine Learning (ICML)}},
  year={2022}
}

@inproceedings{
dino,
title={{DINO}: {DETR} with Improved DeNoising Anchor Boxes for End-to-End Object Detection},
author={Hao Zhang and Feng Li and Shilong Liu and Lei Zhang and Hang Su and Jun Zhu and Lionel Ni and Heung-Yeung Shum},
booktitle = {{The Eleventh International Conference on Learning Representations (ICLR)}},
year={2023}
}

@inproceedings{maskrcnn,
  title={Mask {R-CNN}},
  author={He, Kaiming and Gkioxari, Georgia and Doll{\'a}r, Piotr and Girshick, Ross},
  booktitle = {{Proceedings of the IEEE international conference on computer vision (ICCV)}},
  year={2017}
}

@inproceedings{swin,
  title={Swin transformer: {Hierarchical} vision transformer using shifted windows},
  author={Liu, Ze and Lin, Yutong and Cao, Yue and Hu, Han and Wei, Yixuan and Zhang, Zheng and Lin, Stephen and Guo, Baining},
  booktitle = {{Proceedings of the IEEE/CVF International Conference on Computer Vision (ICCV)}},
  year={2021}
}

@inproceedings{DOTA,
  title={{DOTA}: {A} large-scale dataset for object detection in aerial images},
  author={Xia, Gui-Song and Bai, Xiang and Ding, Jian and Zhu, Zhen and Belongie, Serge and Luo, Jiebo and Datcu, Mihai and Pelillo, Marcello and Zhang, Liangpei},
  booktitle = {{Proceedings of the IEEE/CVF Conference on Computer Vision and Pattern Recognition (CVPR)}},
  year={2018}
}

@inproceedings{AFOG,
  title={Adversarial Attention Perturbations for Large Object Detection Transformers},
  author={Yahn, Zachary and Tekin, Selim Furkan and Ilhan, Fatih and Hu, Sihao and Huang, Tiansheng and Xu, Yichang and Loper, Margaret and Liu, Ling},
  booktitle = {{Proceedings of the IEEE/CVF International Conference on Computer Vision (ICCV)}},
  year={2025}
}

@article{Metric1,
  title={Adversarial metric attack and defense for person re-identification},
  author={Bai, Song and Li, Yingwei and Zhou, Yuyin and Li, Qizhu and Torr, Philip HS},
  journal = {{IEEE Transactions on Pattern Analysis and Machine Intelligence}},
  volume={43},
  number={6},
  pages={2119--2126},
  year={2020},
  publisher={IEEE}
}

@article{Metric3,
  title={Cross-modality perturbation synergy attack for person re-identification},
  author={Gong, Yunpeng and Zhong, Zhun and Qu, Yansong and Luo, Zhiming and Ji, Rongrong and Jiang, Min},
  journal = {{Advances in Neural Information Processing Systems (NeurIPS)}},
  year={2024}
}

@inproceedings{DomainVar,
  title={A robust learning approach to domain adaptive object detection},
  author={Khodabandeh, Mehran and Vahdat, Arash and Ranjbar, Mani and Macready, William G},
  booktitle = {{Proceedings of the IEEE/CVF International Conference on Computer Vision (ICCV)}},
  pages={480--490},
  year={2019}
}

@article{TranferRobust,
  title={Adversarially trained object detector for unsupervised domain adaptation},
  author={Fujii, Kazuma and Kera, Hiroshi and Kawamoto, Kazuhiko},
  journal = {{IEEE Access}},
  volume={10},
  pages={59534--59543},
  year={2022},
  publisher={IEEE}
}

\end{document}